\documentclass{article}

\usepackage{arxiv}

\usepackage[utf8]{inputenc} 
\usepackage[T1]{fontenc}    
\usepackage{hyperref}       
\usepackage{url}            
\usepackage{booktabs}       
\usepackage{amsfonts}       
\usepackage{nicefrac}       
\usepackage{microtype}      
\usepackage{lipsum}		
\usepackage{graphicx}
\usepackage{natbib}
\usepackage{doi}

\usepackage{amsmath}
\usepackage{caption}
\usepackage{subcaption}
\usepackage{pifont}         
\usepackage{threeparttable} 

\graphicspath{{img/}}

\title{DynoDINO: Harnessing Dynamic Latent Information from DINO Features for Multi-Phase Medical Image Segmentation}

\author{
    Yu-Pu Hsu \\
    College of Artificial Intelligence \\
    National Yang Ming Chiao Tung University \\
	Taiwan \\
	\texttt{yupu.ai12@nycu.edu.tw} \\
    \And
	Jen-Jee Chen \\
    College of Artificial Intelligence \\
    National Yang Ming Chiao Tung University \\
    Taiwan \\
	\texttt{jenjee@nycu.edu.tw} \\
    \And
    Yu-Chee Tseng \\
    College of Artificial Intelligence \\
    National Yang Ming Chiao Tung University \\
    Taiwan \\
	\texttt{yctseng@cs.nycu.edu.tw} \\
}

\renewcommand{\shorttitle}{\textit{arXiv} Template}

\hypersetup{
pdftitle={DynoDINO: Harnessing Dynamic Latent Information from DINO Features for Multi-Phase Medical Image Segmentation},
pdfsubject={cs.CV},
pdfauthor={Yu-Pu Hsu, Jen-Jee Chen, Yu-Chee Tseng},
pdfkeywords={Multi-phase contrast-enhanced computed tomography (CECT), Medical image segmentation, Multi-phase fusion, Image alignment, Mix-attention mechanism, Adaptive gating mechanism},
}

\begin{document}
\maketitle

\begin{abstract}
\textbf{Multi-phase Contrast-Enhanced Computed Tomography (CECT)} plays a central role in the diagnosis and characterization of focal lesions by capturing temporal enhancement patterns across multiple acquisition phases. Accurate lesion segmentation from such data remains challenging because clinically relevant contrast kinetics are distributed across phases, while anatomical inconsistencies, respiratory motion, and incomplete acquisitions often lead to inter-phase misalignment and interrupted temporal information. Conventional segmentation frameworks typically process each phase independently or rely on simple fusion strategies, limiting their temporal reasoning capability.

To address these challenges, we propose \textbf{DynoDINO}, a unified framework tailored to address the core challenges of multi-phase medical image segmentation. DynoDINO first performs slice-level alignment to establish inter-phase anatomical correspondence and then employs a \textbf{Multi-phase Fusion Model} to jointly enhance temporal correlations across phases. Our fusion model incorporates a \textbf{Mix-attention (MA)} mechanism for efficient multi-phase feature calibration and an \textbf{Adaptive Gating Mechanism} with difference-based residual learning to selectively preserve diagnostically relevant contrast variations while suppressing artifacts caused by residual misalignment. In addition, the adaptive gating mechanism improves training stability by preventing feature degradation caused by unguided subtraction operations.

Experiments on three large-scale datasets, including \textbf{LiTS}, \textbf{PLC-CECT}, and \textbf{WAW-TACE}, demonstrate that DynoDINO consistently improves boundary delineation and structural fidelity under standard, shifted, and missing-phase conditions. 

\end{abstract}


\section{Introduction}
\label{sec:intro}

Medical imaging plays a fundamental role by supporting disease detection, lesion characterization, treatment planning, and longitudinal patient monitoring \cite{gore2020artificial, esteva2019guide}. 
Accurate lesion segmentation is fundamental to these tasks, as it enables the quantification of lesion size, shape, spatial extent, and treatment response.
Recent advances in deep learning have substantially improved the performance of automated medical image segmentation systems, reducing the workload of radiologists while enabling large-scale image analysis \cite{ronneberger2015u, milletari2016v}. Nevertheless, many clinically important imaging protocols contain rich temporal information that remains underexplored.

\textbf{Contrast-enhanced imaging} is a representative example. In both Computed Tomography (CT) and Magnetic Resonance Imaging (MRI), contrast agents are injected to reveal tissue vascularity and perfusion characteristics over time. As a result, \textbf{multiple imaging phases}, including the \textit{Early Phase (EP)} (arterial phase), \textit{Mid Phase (MP)} (portal venous phase), and \textit{Late Phase (LP)}, are acquired \cite{galle2018easl, chernyak2018liver}. Each phase captures complementary physiological information and provides a different view of the same anatomical structures. The temporal evolution of contrast enhancement, known as \textit{contrast kinetics}, is a key indicator for differentiating benign and malignant lesions \cite{hennedige2013imaging, kamel2005multidetector}. Consequently, radiologists rarely interpret individual phases independently; instead, they rely heavily on comparing enhancement patterns across phases to characterize temporal changes in lesion appearance.


This paper studies multi-phase imaging segmentation of \textbf{Hepatocellular Carcinoma (HCC)}, one of the leading causes of cancer-related mortality  \cite{bray2024global, PracticeGuidance}. Multi-phase contrast-enhanced CT is widely used for HCC diagnosis because characteristic enhancement patterns can often be observed without invasive biopsy. Typical HCC lesions exhibit arterial phase hyper-enhancement followed by relative signal attenuation in later phases \cite{lee2012enhancement, hori2026artificial}. These temporal enhancement characteristics constitute well-established clinical guidelines \cite{ManagementHCC, PracticeGuidance, TreatmentGuidelines}. From a segmentation perspective, lesion boundaries may become more distinguishable in certain phases, while other phases provide complementary contextual information regarding vascular structure, lesion extent, and surrounding tissue characteristics. 

Despite the importance of multi-phase imaging in clinical diagnosis, most existing segmentation frameworks are fundamentally tailored to single-phase image analysis.
When multiple phases are available, they are typically incorporated through channel-wise concatenation \cite{isensee2018nnu} or other simple fusion strategies \cite{zhou2019review}. However, these approaches often dilute cross-phase interactions and fail to capture diagnostically important temporal enhancement patterns.
More advanced transformer-based architectures have demonstrated strong capabilities in modeling long-range dependencies \cite{dosovitskiy2020image, hatamizadeh2022unetr}. Nevertheless, directly applying these models to multi-phase imaging remains challenging due to three practical issues commonly encountered in clinical datasets.
First, multi-phase acquisitions often suffer from \textbf{inter-slice misalignment}. Since different phases are acquired at separate time points, anatomical correspondence along the z-axis is frequently disrupted by respiratory motion, patient movement, and acquisition variability, making reliable slice-to-slice correspondence difficult to establish \cite{haskins2020deep}.
Second, clinically relevant enhancement patterns are inherently \textbf{non-linear}. Effective lesion characterization therefore requires modeling temporal contrast dynamics rather than simply aggregating multi-phase features.
Third, real-world multi-phase imaging is often incomplete, with \textbf{missing phases} caused by variations in clinical imaging protocols. However, existing segmentation frameworks rarely account for such practical scenarios.
Collectively, these challenges call for a fundamental redesign of existing segmentation frameworks.

Recent advances in self-supervised learning and medical foundation models have demonstrated the potential of pre-trained representations for medical image analysis \cite{caron2021emerging, meddinov3}. However, effective utilization of multi-phase temporal information with these pretraining models remains an open problem. In particular, the combined challenges of multi-phase alignment, temporal non-linearity, and robustness to incomplete phases need to be addressed.
To address these limitations, we propose \textbf{DynoDINO}, a general-purpose end-to-end framework for multi-phase medical image segmentation that integrates inter-phase alignment, temporal representation learning, adaptive feature fusion, and robust lesion segmentation. 

The main contributions of this work are summarized as follows:



\begin{itemize}
    \item \textbf{Robust Multi-Phase Architecture:} We propose a multi-phase architecture with a \textbf{Flexible Input} that natively accommodates incomplete phase sequences, enabling robust learning from heterogeneous clinical acquisition protocols. The architecture integrates an \textbf{FFT-based ZNCC alignment} and an \textbf{Adaptive Gating Module} to effectively suppress registration noise and subtraction artifacts, preventing representation collapse caused by missing phases and stabilizing the difference-based learning.
       
    \item \textbf{Efficient Mix-Attention (MA):} We introduce an MA mechanism for efficient \textbf{Cross-Phase Fusion}. By leveraging a shared bidirectional attention map, the model effectively captures complementary temporal phase dynamics while avoiding redundant attention computation. Furthermore, we incorporate \textbf{FlashAttention} \cite{dao2022flashattention} to alleviate GPU memory bottlenecks. Compared with conventional cross-attention, DynoDINO reduces the memory footprint by 4.1\%, corresponding to a savings of approximately 7.08M parameters.
       
    \item \textbf{Unified Framework:} We establish a unified framework for multi-phase medical image segmentation that seamlessly integrates alignment, temporal modeling, and adaptive fusion into a single workflow. 
    Although evaluated on multi-phase contrast-enhanced CT datasets, the proposed architecture is readily extensible to other multi-phase or multi-temporal medical imaging modalities without requiring protocol-specific redesign.
    
    \item \textbf{Extensive Clinical Validation:} Comprehensive experiments across three large-scale datasets with distinct clinical protocols (\textbf{LiTS}, \textbf{PLC-CECT}, and \textbf{WAW-TACE}) demonstrate consistent improvements in boundary delineation and structural fidelity under standard, shifted, and missing-phase conditions.

\end{itemize}

\section{Related Work}
\label{sec:related_work}


\textbf{DINO Family.}
Recent advances in self-supervised learning have led to powerful visual foundation models. The DINO family \cite{caron2021emerging}, including DINOv2 \cite{oquab2023dinov2}, DINOv3 \cite{simeoni2025dinov3}, and \textbf{MedDINOv3} \cite{meddinov3}, has demonstrated strong performance across general and medical vision tasks. For dense prediction, \textbf{Dino U-Net} \cite{gao2025dino} integrates DINO features with a U-Net architecture for volumetric segmentation. Recent studies \cite{zhang2023multimodal} further suggest that foundational representations can provide richer semantic information than conventional backbones such as \textbf{nnU-Net} \cite{isensee2018nnu} and \textbf{SegFormer} \cite{xie2021segformer}. Despite these advances, existing DINO-based segmentation frameworks are primarily designed for single-phase image analysis, leaving multi-phase imaging largely unexplored.

\textbf{Image Registration and Alignment.}
Traditional registration methods align images by optimizing similarity measures such as Mutual Information and Cross-Correlation \cite{zitova2003image}. More recently, learning-based approaches such as \textbf{VoxelMorph} \cite{balakrishnan2019voxelmorph} estimate deformation fields directly from image pairs. However, deformable registration remains computationally expensive for high-resolution volumetric data, while large inter-phase contrast variations can introduce registration errors and artifacts. Consequently, several studies \cite{zhang2021modality, estienne2019u} have explored segmentation frameworks that do not require precise voxel-level alignment. These limitations motivate a framework that tolerates residual inter-phase inconsistencies while effectively exploiting complementary information across phases.




\textbf{Multi-phase Image Fusion.}
Different fusion strategies have been studied for multi-phase medical image segmentation. Early fusion approaches \cite{isensee2018nnu, zhou2019review} combine data at the input level by channel concatenation. Late fusion methods process each phase independently and merge high-level representations at deeper network stages \cite{zhou2019review, bi2021recurrent}. More recently, attention-based fusion techniques \cite{liu2022transfusion, papanastasiou2023attention} have been introduced to adaptively aggregate information.
Despite these advances, temporal enhancement patterns are typically represented implicitly through feature aggregation rather than modeled explicitly. However, in clinical practice, lesion characterization relies heavily on analyzing enhancement changes across phases.




\textbf{Temporal Representation Learning.}
Multi-phase CT can be viewed as a sparse temporal sequence. Temporal representation learning has been extensively studied in video understanding \cite{wang2018non, feichtenhofer2019slowfast} and change detection \cite{chen2021remote}, where difference representations emphasize structural changes across time. Similarly, clinical lesion characterization relies on arterial enhancement and delayed wash-out patterns observed across phases. However, explicit modeling of inter-phase enhancement dynamics remains uncommon in current 3D medical image segmentation frameworks.



\textbf{Efficient Attention Mechanisms.}
The quadratic complexity of self-attention limits the scalability of Transformer-based models for volumetric imaging. Efficient attention mechanisms \cite{zhou2021informer, tay2022efficient} and recent advances such as \textbf{Flash Attention} \cite{dao2023flashattention} reduce computational and memory costs while preserving long-range dependency modeling. These techniques are particularly beneficial for multi-phase volumetric imaging, where large feature maps impose substantial memory demands.


\section{Method}
\label{sec:method}

\subsection{Multi-Phase Imaging Datasets}

The appearance of lesions in CECT scans varies substantially across different phases. Pre-contrast phases often lack the structural detail necessary for precise lesion characterization and are frequently absent from clinical datasets. Consequently, our framework focuses on the three primary post-contrast phases. To ensure clinical generalizability, we define a generalized temporal hierarchy. Despite variations in organ-specific nomenclature (e.g., the corticomedullary phase in renal imaging), we categorize the sequences into Early Phase (EP), Mid Phase (MP), and Late Phase (LP). These stages represent a generalized temporal hierarchy: the initial arrival of the contrast agent, the peak parenchymal enhancement, and the subsequent clearance or equilibrium stage, respectively. This taxonomy ensures that DynoDINO remains applicable to various contrast-enhanced modalities, including multi-parametric MRI (mpMRI) and dynamic CT protocols beyond liver imaging.

In this work, we consider three multi-phase datasets:
\begin{itemize}
    \item \textbf{LiTS (Liver Tumor Segmentation)}: Sourced from the MSD Challenge (Task03\_Liver) \cite{bilic2023liver}, this is a single-phase dataset. It consists of 201 3D volumes, with 131 patients provided with ground-truth labels for the liver and associated tumors. LiTS serves as a single-phase benchmark in our experiments. We adapt these volumes into our multi-phase framework by routing all frames to the EP channel and all missing frames to the MP and LP channels (intensity value $= -1024$ HU), thereby requiring no architecture change.
    
    \item \textbf{PLC-CECT (Primary Liver Cancer CECT Imaging Dataset)}: Sourced from Science Data Bank \cite{plccect}, this is our core multi-phase dataset providing comprehensive temporal enhancement profiles. It includes 278 patients presenting with HCC, Intrahepatic Cholangiocarcinoma (ICC), and Combined HCC-CCA (cHCC-CCA), along with 83 control subjects without liver cancer. A total of 50,560 lesion-bearing 2D slices were collected.
    
    \item \textbf{WAW-TACE}: Sourced from Zenodo \cite{bartnik2024waw}, this dataset represents a real-world challenge characterized by incomplete temporal sequences. Since Transarterial Chemoembolization (TACE) is the primary first-line treatment for HCC patients unsuitable for surgery, this dataset includes data from 233 treatment-naive HCC patients undergoing TACE. It contains 377 independent labels, where multiple tumors within the same patient are treated as distinct labeled entities, reflecting the complex multifocal nature of liver cancer in clinical practice.
\end{itemize}

\textbf{Exclusion Criteria.} 
To ensure the integrity of multi-phase fusion training, the following data filtering criteria were applied:
\begin{enumerate}
    \item \textbf{Pre-contrast Exclusion:} Non-contrast phases were excluded due to their limited structural information.
    \item \textbf{EP Requirement:} Patients without the critical EP were excluded to ensure the presence of the baseline enhancement signature.
    \item \textbf{Annotation Requirement:} Patient volumes without ground-truth annotations were excluded.
\end{enumerate}

\begin{table}[t]
\centering
\footnotesize 
\begin{threeparttable}
\caption{Distribution of the datasets after filtering.}
\label{tab:dataset_dist}
\begin{tabular}{lccc}
\toprule
\textbf{Category} & \textbf{LiTS} & \textbf{PLC-CECT} & \textbf{WAW-TACE} \\
\midrule
Patients & 131 & 361 & 123 \\
2D Slices & 58,638 & 46,761 & 29,037 \\
\midrule
(1, 1, 1) & 0 & 46,761 & 18,951 \\
(1, 0, 1) & 0 & 0 & 2,720 \\
(1, 1, 0) & 0 & 0 & 9,074 \\
(1, 0, 0) & 58,638 & 0 & 1,708 \\
\bottomrule
\end{tabular}
\begin{tablenotes}[para,flushleft]
\scriptsize
\item \textit{Note: The binary triplet denotes the existence EP, MP, and LP phases, where 1/0 indicates the presence/absence of the respective phase (e.g., (1, 0, 1) represents the absence of MP).}
\end{tablenotes}
\end{threeparttable}
\end{table}

DynoDINO integrates multi-phase features to enhance representation learning, but generates a single, unified segmentation mask. To maintain spatial alignment, the final prediction is mapped directly to the EP reference frame.

\subsection{Data Preprocessing with FFT-based ZNCC}

Multi-phase CECT datasets are often unstandardized.
During scanning across different phases, patient respiration or involuntary movement can lead to anatomical misalignment, as illustrated in Fig.~\ref{fig:alignment_viz}(a). Furthermore, inconsistent slice thicknesses between sequences pose a significant challenge for precise pixel-wise fusion. While traditional methods rely on 3D geometric transformations, they often struggle with the drastic intensity shifts inherent in contrast-enhanced imaging.

To address this issue, we employ \textbf{FFT-accelerated Zero-mean Normalized Cross-Correlation (ZNCC)} for 2D slice alignment. Its robustness to linear brightness and contrast variations \cite{lewis1995fast} makes it well suited for multi-phase CECT with substantial inter-phase intensity variations.
We then select one of EP, MP, or LP as the \textit{Fixed Phase (FP)} to serve as the alignment reference using the following strategy, while the remaining unselected phases are designated as the \textit{Source Phases (SP)}.
\begin{itemize}
    \item \textbf{Complete Sequences:}
    For datasets (e.g., PLC-CECT) in which all temporal phases are available but slice counts differ due to varying reconstruction intervals, the phase with the fewest slices is selected as the FP to preserve data integrity and minimize interpolation.
    \item \textbf{Incomplete Sequences:}
    For datasets (e.g., WAW-TACE) with missing phases but a guaranteed baseline, EP is selected as the FP to ensure a consistent anatomical reference across all subjects.
\end{itemize}

\textbf{Spatial Resizing.}
Prior to ZNCC calculation, all 2D slices are resized to a uniform resolution of $512 \times 512$ pixels, ensuring identical matrix dimensions required for FFT operations.

\textbf{ZNCC Scoring.}
The alignment is performed on a per-patient basis. Based on the above strategy, the FP is established, while the remaining phases serve as SP. For each slice in the FP, we iterate through all slices in the SP to calculate ZNCC scores. The index corresponding to the peak score is identified as the best match candidate.

\textbf{Reliability Verification.}
To prevent false matching—particularly in regions with sparse anatomical structures (e.g., empty abdominal space where multiple slices appear similar)—we perform the following verification protocol:
\begin{enumerate}
    \item Thresholding: 
    The peak ZNCC similarity score must exceed a predefined similarity threshold (0.75).
    \item Statistical Significance:
    The peak score must be sufficiently separated from the remaining score distribution. Specifically, the difference between the maximum score and the median score must be greater than 1.5 times the difference between the 75th percentile and the median.
\end{enumerate}
If a reliable match is not found (e.g., the anatomy is outside the scan range), the slice is padded with a default value ($-1024$ HU for CT, 0 for labels), which effectively discards unaligned frames to minimize noise interference. Otherwise, the slices are sequentially concatenated to the whole volume as detailed below.

\textbf{FFT-based Acceleration.}
Calculating ZNCC for every slice pair is computationally expensive. We leverage the Fast Fourier Transform (FFT) to accelerate this process. Mathematically, convolution in the spatial domain is equivalent to element-wise multiplication in the frequency domain \cite{guizar2008efficient}. By utilizing GPU-accelerated FFT, we can significantly reduce the computational overhead. Without loss of generality, we consider a FP slice $I_{\text{fixed}}$ and a SP slice $I_{\text{source}}$. In practice, multiple SP slices can be evaluated in parallel as a batch on a GPU for parallelism.

It includes 4 steps:

\noindent
\textbf{A. Mean Removal.}
To remove global intensity offsets and improve robustness across scanning devices, each slice is first mean-centered:
\begin{equation}
    \hat{I}_{\text{fixed}} = I_{\text{fixed}} - \mu_{I_{\text{fixed}}}, \quad \hat{I}_{\text{source}} = I_{\text{source}} - \mu_{I_{\text{source}}}
\end{equation}
where $\mu_{I_{\text{fixed}}}$ and $\mu_{I_{\text{source}}}$ represent the scalar spatial means of the FP and SP slices, respectively.

\noindent
\textbf{B. Cross-Correlation in the Frequency Domain.}
The ZNCC numerator $N$ is computed using the convolution theorem:
\begin{equation}
    N = \mathcal{F}^{-1} (\mathcal{F}(\hat{I}_{\text{source}}) \odot \mathcal{F}(\hat{I}_{\text{fixed}})^*)
\end{equation}
where $\mathcal{F}$ and $\mathcal{F}^{-1}$ denote the 2D FFT and inverse FFT, respectively, and $\mathcal{F}(\hat{I}_{\text{fixed}})^*$ denotes the complex conjugate of the FP slice spectrum.

\noindent
\textbf{C. Local Normalization Factor.}
The ZNCC denominator $D$, representing the local energy, is similarly accelerated via FFT:
\begin{equation}
    D = \sqrt{\mathcal{F}^{-1} (\mathcal{F}(\hat{I}_{\text{source}}^2) \odot \mathcal{F}(\mathbf{1})^*) \odot \|\hat{I}_{\text{fixed}}\|_2^2 + \epsilon}
\end{equation}
where $\hat{I}_{\text{source}}^2$ is the element-wise square of the mean-centered source batch, $\mathbf{1}$ is a kernel of ones of the FP slice size, and $\|\hat{I}_{\text{fixed}}\|_2^2$ is the total energy of the mean-centered fixed slice (corresponding to \texttt{T\_std\_sum\_sq} in the implementation). $\epsilon$ is a smoothing term ($10^{-8}$) to avoid division by zero.

\noindent
\textbf{D. Final ZNCC Score.}
The normalized correlation matrix is defined as:
\begin{equation}
    ZNCC(I_{\text{source}}, I_{\text{fixed}}) = N/D
\end{equation}
During input, we select the $I_{\text{source}}$ with the highest ZNCC value as the best-matched reference SP slice for each $I_{\text{fixed}}$.

\textbf{Slice Alignment of the Input Volume.} 
The final 3D input volume is synthesized through FP-to-SP re-indexing. Specifically, for each FP slice $I_{\text{fixed}}$, the best-matched SP slice $I_{\text{source}}$ from each of the other two phases is selected and aligned with the FP slice in the input volume, ensuring improved anatomical correspondence across phases.

\begin{figure*} 
    \centering
    \includegraphics[width=\linewidth]{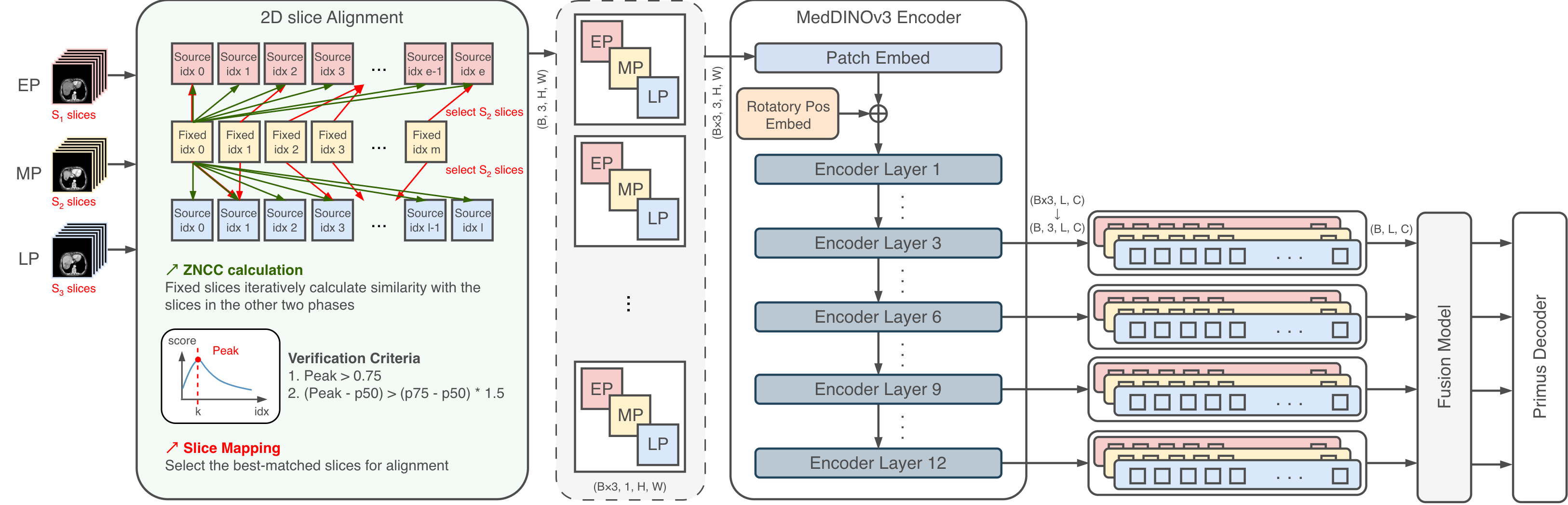}
\caption{The DynoDINO framework. Multi-phase CT slices are first aligned using the ZNCC matching method (here, the MP is assumed to serve as the FP). The aligned slices are reshaped from $(B, 3, H, W)$ to $(B\times 3, 1, H, W)$ and then replicated to $(B \times 3, 3, H, W)$ to comply with the MedDINOv3 encoder input format. After phase-wise feature extraction, the encoded representations are reshaped from $(B \times 3, L, C)$ back to $(B, 3, L, C)$ for cross-phase fusion. Finally, the fused representation $(B, L, C)$ is decoded by the Primus Decoder.}
    \label{fig:overall_arch}
\end{figure*}

\begin{figure}[t]
    \centering
    \includegraphics[width=0.4\linewidth]{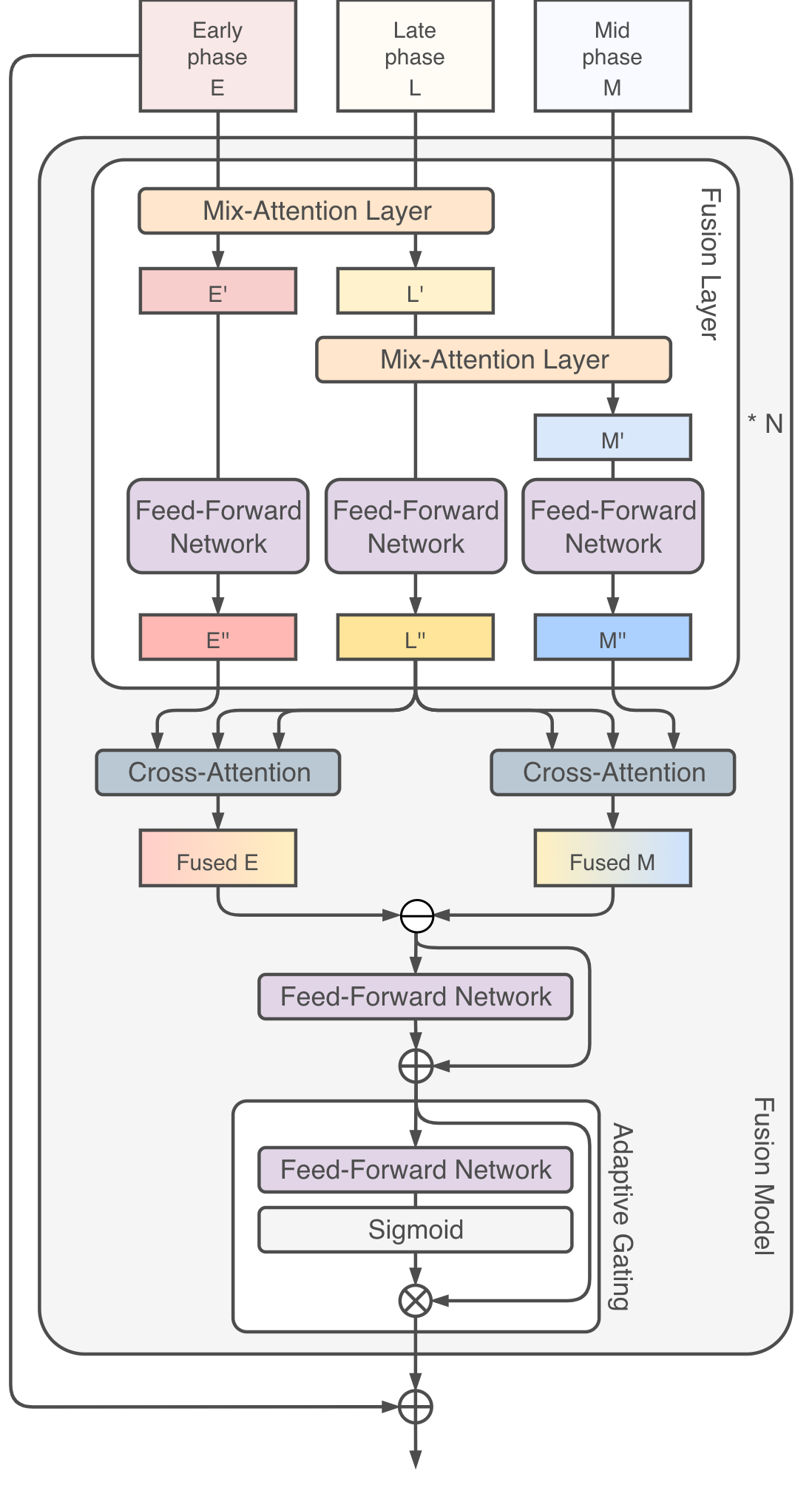}
    \caption{The Fusion Module, which performs $N$ iterations of Mix-Attention followed by a difference-based residual connection and a Sigmoid gating mechanism.}
    \label{fig:fusion_model}
\end{figure}

\begin{figure}[t]
    \centering
    \includegraphics[width=0.4\linewidth]{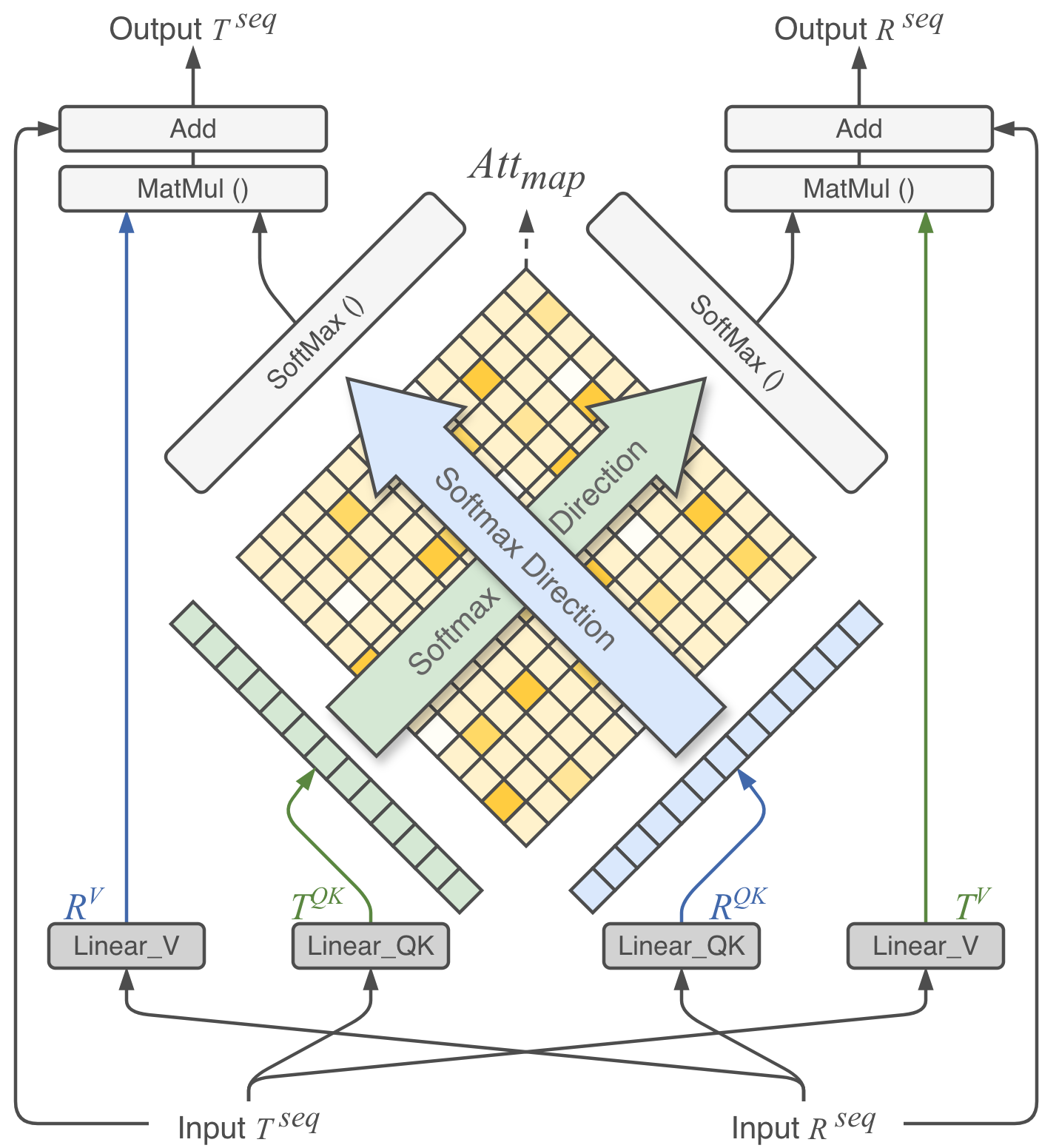}
\caption{The Mix-Attention Layer computes a shared bidirectional attention map ($Att_{map}$) between the reference LP (input $T$) and the target EP/MP (input $R$), reducing computational overhead while effectively capturing dynamic contrast boundaries.}
    \label{fig:mix_attention_layer}
\end{figure}

\subsection{DynoDINO Architecture}

We propose the DynoDINO framework, illustrated in Fig.~\ref{fig:overall_arch}. It adopts MedDINOv3 \cite{meddinov3}, a medical adaptation of DINOv3 \cite{simeoni2025dinov3}, as the backbone encoder. A novel \textbf{Fusion Module} is introduced to fuse the representations from multiple physiological phases (EP, MP, and LP). The module includes iterative \textbf{Fusion Layers}, followed by \textbf{Difference-based Residual} and \textbf{Adaptive Gating}.

\subsubsection{Input Reshaping and Multiscale Extraction}

To facilitate parallel processing, 2D slices from the three phases are concatenated along the batch dimension. In instances of missing phases (as observed in WAW-TACE), inputs are padded with a background intensity value of $-1024$ HU. Through the MedDINOv3 encoder, multiscale latent representations are extracted from layers 3, 6, 9, and 12. This hierarchical extraction captures a wide range of receptive fields, including both global anatomical structures and localized, subtle focal lesions.

\subsubsection{Fusion Layers and Mix-Attention}

While MedDINOv3 encodes each phase independently, the $N$ Fusion Layers (e.g., $4$) then progressively aligns and calibrates features across temporal phases, as illustrated in Fig.~\ref{fig:fusion_model}.

\textbf{Mix-Attention (MA) Mechanism.}
Typical attention for high-resolution multi-phase CECT data is computationally expensive.
The MA is designed to balance attention effectiveness and efficiency. Unlike traditional cross-attention that computes separate maps for each directional interaction, our MA utilizes a \textbf{shared} attention map. A single affinity matrix $\mathbf{Att}_{map}$ is computed bidirectionally between the reference (LP) and target (EP/MP) streams. The design is inspired by efficient transformer variant Informer \cite{zhou2021informer}, which utilizes shared query-key structures to reduce redundant computations. As demonstrated in our complexity analysis (Table~\ref{tab:ablation}), this strategy reduces trainable projection parameters and activation memory, achieving a 4.1\% reduction (effectively 7.08M less) in total parameters compared to standard cross-attention baselines. Combined with \textbf{Flash Attention} \cite{dao2022flashattention}, it enables DynoDINO to process multi-scale sequences without GPU memory bottlenecks.

\textbf{Mathematical Formulation.}
Let $\mathbf{T}^{seq}$ (LP) and $\mathbf{R}^{seq}$ (EP/MP) denote the latent features. The projections are defined as:
\begin{align}
    \mathbf{R}^{QK}, \mathbf{T}^{QK}, \mathbf{R}^{V}, \mathbf{T}^{V} = \text{Linear}(\cdot)
\end{align}
The shared attention matrix $\mathbf{Att}_{map}$ captures inter-phase affinity:
\begin{equation}
    \mathbf{Att}_{map} = \mathbf{R}^{QK} \times \mathbf{T}^{QK}
\end{equation}
The calibrated outputs are derived by applying the map in opposing directions:
\begin{align}
    \mathbf{R}'^{seq} &= \mathbf{R}^{QK} + \text{Softmax}\left(\frac{\mathbf{Att}_{map}}{\sqrt{d_k}}\right) \mathbf{T}^{V} \\
    \mathbf{T}'^{seq} &= \mathbf{T}^{QK} + \text{Softmax}\left(\frac{\mathbf{Att}_{map}^T}{\sqrt{d_k}}\right) \mathbf{R}^{V}
\end{align}

\subsubsection{Difference-based Residual and Adaptive Gating}

After $N$ Fusion Layers, we obtain the refined multi-phase features $\mathbf{E}''$, $\mathbf{L}''$, and $\mathbf{M}''$. Then, they are passed through a pair of post-loop Cross-Attention Layers anchored by the LP representation to generate the final aligned reference streams, denoted as $\mathbf{F}_E$ (Fused E) and $\mathbf{F}_M$ (Fused M):
\begin{align}
    \mathbf{F}_E &= \text{Cross-Att}(\mathbf{Q}=\mathbf{E}'', \mathbf{K}=\mathbf{L}'', \mathbf{V}=\mathbf{L}'') \\
    \mathbf{F}_M &= \text{Cross-Att}(\mathbf{Q}=\mathbf{M}'', \mathbf{K}=\mathbf{L}'', \mathbf{V}=\mathbf{L}'')
\end{align}

Building upon these fully calibrated representations, the Difference-based Residual Learning aims to mimic the \textbf{clinical subtraction logic}. To ensure numerical stability before subtraction, layer normalization ($\text{LN}$) is applied:
\begin{align}
    \Delta \mathbf{F}_{diff} = \text{LN}(\mathbf{F}_E) - \text{LN}(\mathbf{F}_M) \\
    \Delta \mathbf{F} = \Delta \mathbf{F}_{diff} + \text{FFN}_{feat}(\Delta \mathbf{F}_{diff})
\end{align}
Then the Adaptive Gating selectively incorporates these temporal variances with a goal of \textbf{suppressing alignment noise}. The residual difference feature is normalized again to generate a spatial weight map $\mathbf{W}$:
\begin{equation}
    \mathbf{W} = \sigma(\text{FFN}_{gate}(\text{LN}(\Delta \mathbf{F})))
\end{equation}
The final output $\mathbf{F}_{out}$ is constructed by injecting the gated difference back into the original EP features $\mathbf{E}_{in}$ (which corresponds to the pre-fusion representation):
\begin{equation}
    \mathbf{F}_{out} = \mathbf{E}_{in} + (\Delta \mathbf{F} \odot \mathbf{W})
\end{equation}
Clinically, this gating mechanism preserves sensitivity to phase-specific enhancement patterns, such as wash-in and wash-out kinetics, while remaining robust to residual local misalignment.

\subsection{Compound Loss Function}

To ensure that the DynoDINO framework remains robust across diverse clinical scenarios and anatomical structures, we employ a compound loss function $\mathcal{L}_{total}$. While our focus is liver and tumor segmentation, it follows the MedDINOv3 \cite{meddinov3} implementation to address the universal challenges of medical imaging: high class imbalance and the need for precise boundary localization.
\begin{equation}
    \mathcal{L}_{total} = \mathcal{L}_{Dice} + \mathcal{L}_{CE}
\end{equation}

\textbf{Pixel-wise Cross-Entropy Loss.}
The Cross-Entropy component $\mathcal{L}_{CE}$ treats segmentation as a comprehensive pixel-wise classification task. By optimizing the log-likelihood of each class, it provides a stable and smooth gradient landscape, which is essential for the convergence of the Vision Transformer backbone during the initial training phases. For $N$ pixels and $C$ classes, it is defined as:
\begin{equation}
    \mathcal{L}_{CE} = -\frac{1}{N} \sum_{i=1}^{N} \sum_{c=1}^{C} y_{i,c} \log(\hat{y}_{i,c})
\end{equation}
where $y_{i,c}$ is the ground-truth label and $\hat{y}_{i,c}$ is the predicted probability for class $c$ at pixel $i$.

\textbf{Region-based Soft Dice Loss.}
To mitigate the impact of class imbalance—a common issue in medical imaging where the target organs or lesions often occupy a small fraction of the total volume—we incorporate the Soft Dice Loss. This region-based component directly optimizes the overlap between the prediction and ground truth, ensuring that the model remains sensitive to small structural details regardless of the specific organ type.
\begin{equation}
    \mathcal{L}_{Dice} = 1 - \frac{2 \sum_{i=1}^{N} \hat{y}_{i,c} y_{i,c} + \epsilon}{\sum_{i=1}^{N} \hat{y}_{i,c} + \sum_{i=1}^{N} y_{i,c} + \epsilon}
\end{equation}
where $\epsilon = 10^{-5}$ is a smoothing constant for numerical stability.

\section{Experiments}
\label{sec:results}

\subsection{Implementation Details}

We performed experiments using PyTorch and NVIDIA RTX 4090 GPUs. To ensure a fair and transparent comparison, all experiments were conducted within the official MedDINOv3 implementation settings \cite{meddinov3}. Unless otherwise specified, we strictly adhered to the default hyperparameters provided by the MedDINOv3 repository:
\begin{itemize}
    \item \textbf{Optimization}: AdamW optimizer (Weight Decay $= 0.05$) with a SequentialLR scheduler (10-epoch linear warmup, 90-epoch cosine annealing).
    \item \textbf{Loss Function}: A hybrid of Soft Dice Loss and Cross-Entropy (CE) Loss.
    \item \textbf{Preprocessing}: Standardized intensity clipping and resampling consistent with the MedDINOv3 data pipeline.
\end{itemize}
This protocol ensures that the observed performance gains are directly attributable to the architectural innovations of DynoDINO.

For datasets with missing phases (specifically the WAW-TACE cohort), standard configurations of CNN-based baselines like nnU-Net inherently crash due to mismatched input channels. To enable these baselines to execute without runtime errors, a preprocessing protocol was applied: any missing phase was artificially instantiated by blank volumes, with the slice count perfectly matched to the existing reference phases. Consequently, all models were evaluated on structurally consistent inputs.


\subsection{Evaluation Metrics and Baselines}

Three standard metrics were evaluated: Dice Similarity Coefficient (DSC), Normalized Surface Distance (NSD), and the 95th percentile Hausdorff Distance (HD95) \cite{taha2015metrics}. 

While DSC provides a global evaluation of regional overlap, it is inherently volume-biased and relatively insensitive to localized boundary errors. In clinical practice—especially for focal hepatic lesion characterization and surgical resection planning—\textbf{boundary precision} is far more critical than generic region overlap. Misalignments at the tumor margins directly impact the determination of safe surgical borders and the evaluation of residual viable tumor patches. Therefore, NSD and HD95 were heavily prioritized in this study to evaluate the morphological fidelity across phases.

For multi-phase datasets, we evaluated DynoDINO under two configurations:
\begin{itemize}
    \item \textbf{Single-phase Input}: Only the EP is provided; MP and LP are padded with background values ($-1024$ HU).
    \item \textbf{Multi-phase Input}: All available phases (EP, MP, LP) are utilized. 
\end{itemize}
We benchmarked DynoDINO against CNN-based nnU-Net \cite{isensee2018nnu}, SegFormer \cite{xie2021segformer}, Dino U-Net \cite{gao2025dino}, and vanilla MedDINOv3 \cite{meddinov3}.

\subsection{Benchmark Results}

\subsubsection{Robustness under Single-phase Constraints}

We evaluate on LiTS (Tables~\ref{tab:lits_results}--\ref{tab:lits_spatial}) and under single-phase constraints on LPC-CECT and WAW-TACE (Tables~\ref{tab:plc_cect_results}--\ref{tab:waw_tace_spatial}).
These results demonstrate the inherent robustness of DynoDINO to incomplete phase information.
In LiTS (Table~\ref{tab:lits_results}), DynoDINO achieves a superior global HD95 of 13.76 mm compared to MedDINOv3 (13.85 mm), suggesting that even without temporal cues, the self-supervised spatial priors from our backbone coupled with the Iterative Refinement module provide stronger structural regularization.

This structural integrity is further verified by the spatial localization metrics in Table~\ref{tab:lits_spatial}. DynoDINO yields a highly precise tumor centroid distance (CD) of 7.56 mm, markedly outperforming the gold-standard nnU-Net (11.26 mm). It is worth noting that DynoDINO exhibits a larger average minimum detectable tumor volume ($\text{Avg.\ } V_{\text{min}}$) of 4,826.23 $\text{mm}^3$, a limitation inherited from Vision Transformer (ViT) architectures due to patch-level tokenization that inherently sacrifices pixel-level local resolution. Nevertheless, the substantial reduction in tumor CD demonstrates that DynoDINO successfully constrains severe spatial deviations and prevents topological hallucinations during voxel clustering.

\subsubsection{Performance under Multi-phase Settings}

It is worth noting that although foundation models (e.g., Dino U-Net) theoretically possess more representative latent embeddings, standard nnU-Net often yields superior or competitive Dice scores in single-phase or naive multi-phase settings. This discrepancy is primarily attributed to nnU-Net's heavily optimized preprocessing pipeline and aggressive data augmentation. However, when confronting complex temporal contrast kinetics, both architectures fall short due to the lack of dedicated inter-phase modeling, a limitation effectively addressed by DynoDINO.

The clinical effectiveness of this explicit multi-phase temporal modeling is clearly reflected in the results. 
In the clinical PLC-CECT dataset (Table~\ref{tab:plc_cect_results}), transitioning from single-phase to multi-phase inputs allows DynoDINO to significantly reduce the global HD95 from 26.81 mm to 24.35 mm. Conversely, while nnU-Net maintains a marginally higher Global DSC (79.34\% vs. 78.00\%), this regional advantage is fundamentally driven by the CNN's inherent \textbf{volume-matching and over-smoothing bias}. 
In our qualitative study (Sec.~\ref{sec:qual-analysis}),
nnU-Net achieves high region overlap by generating generalized, blob-like masks that sacrifice complex structural margins, leading to a drastically inferior HD95 (28.72 mm). 

DynoDINO's explicit modeling of inter-phase kinetics and self-supervised structural priors strictly prioritizes edge precision. Under the multi-phase configuration, DynoDINO offers a 15.2\% reduction in global HD95 (24.35 mm vs. 28.72 mm) and a superior global NSD (47.10\%) compared to its multi-phase nnU-Net counterpart. 

Furthermore, our multi-phase alignment empowers DynoDINO to achieve a remarkable leap in small-lesion detectability, reducing the liver $\text{Avg.\ } V_{\text{min}}$ from 90,145.65 $\text{mm}^3$ down to an exceptional 21,871.95 $\text{mm}^3$
(Table~\ref{tab:plc_cect_spatial}). These findings demonstrate that DynoDINO successfully captures true physiological contrast kinetics rather than relying on over-smoothed region approximations.

\subsubsection{Robustness to Real-world Data Incompleteness}

The WAW-TACE dataset represents a highly challenging post-treatment scenario characterized by severe artifacts and \textbf{inconsistent scanning intervals}. Counter-intuitively, for the CNN baselines such as  nnU-Net, transiting from a single-phase to a multi-phase configuration leads to a systematic degradation: the macro-averaged Global DSC drops from 79.04\% to 77.75\% and NSD drops from 54.19\% to 53.11\
 (Table~\ref{tab:waw_tace_results}).

This performance drop is directly tied to the artificial imputation protocol required to prevent pipeline execution failure. Because standard nnU-Net lacks an intrinsic routing mechanism to handle missing temporal sequences, the brute-force padding ($-1024$ HU) introduces severe non-physiological step-artifacts. Consequently, the localized convolutional layers are forced to ingest contaminated features, disrupting the network's spatial consistency.

Operating in the multi-phase domain, DynoDINO remains effectively insulated from the imputation noise.
While the single-phase nnU-Net achieves a high surface alignment (NSD of 54.19\%) due to its highly optimized voxel-intensity preprocessing on complete reference volumes, it exhibits a dangerous vulnerability to localized boundary anomalies under severe lipiodol artifacts, as reflected by its inferior HD95 (21.31 mm). Conversely, driven by the Adaptive Gating mechanism, DynoDINO actively filters the padded placeholders, recovering an outstanding global NSD (54.69\%), surpassing the single-phase CNN, while simultaneously achieving a substantial 13.0\% reduction in global HD95 (18.54 mm) compared to the single-phase nnU-Net baseline. 

DynoDINO also achieves the lowest liver centroid distance (1.61 mm) under extreme imaging noise (Table~\ref{tab:waw_tace_spatial}). 
These results demonstrate that DynoDINO effectively mitigates the topological hallucinations and missed lesions commonly observed in vanilla architectures, thereby ensuring high boundary precision for real-world applications.


\begin{table*}[t]
\centering
\caption{Quantitative segmentation performance on the LiTS dataset. 
We report organ-, lesion-level, and global segmentation sensitivity.}
\label{tab:lits_results}
\footnotesize
\setlength{\tabcolsep}{5pt}
\begin{tabular}{l cccccc c}
\toprule
 & \multicolumn{2}{c}{\textbf{Liver}} & \multicolumn{2}{c}{\textbf{Tumor}} & \multicolumn{3}{c}{\textbf{Global Performance}} \\
\cmidrule(r){2-3} \cmidrule(lr){4-5} \cmidrule(l){6-8}
Method & DSC (\%) $\uparrow$ & Sens. (\%) $\uparrow$ & DSC (\%) $\uparrow$ & Sens. (\%) $\uparrow$ & DSC (\%) $\uparrow$ & NSD (\%) $\uparrow$ & HD95 (mm) $\downarrow$ \\
\midrule
SegFormer       & 94.06$\pm$3.18 & 94.70$\pm$2.00 & 52.88$\pm$18.82 & 57.08$\pm$35.45 & 73.47 & 44.24 & 26.79 \\
Dino U-Net      & 94.25$\pm$3.03 & 95.22$\pm$2.04 & 55.79$\pm$20.00 & 64.19$\pm$17.36 & 75.02 & 45.39 & 37.03 \\
nnU-Net         & 95.19$\pm$5.04 & 96.33$\pm$1.54 & 58.74$\pm$19.37 & \textbf{68.11}$\pm$15.36 & 76.97 & 49.23 & 19.01 \\
MedDINOv3       & \textbf{96.33}$\pm$1.07 & 96.53$\pm$1.28 & 56.82$\pm$18.22 & 60.30$\pm$16.58 & 76.58 & \textbf{52.31} & 13.85 \\
\textbf{DynoDINO (ours)} & 96.16$\pm$0.57 & \textbf{97.02}$\pm$1.18 & \textbf{60.12}$\pm$15.57 & 55.81$\pm$15.32 & \textbf{78.14} & 49.19 & \textbf{13.76} \\
\bottomrule
\end{tabular}
\end{table*}

\begin{table*}[t]
\centering
\caption{Spatial localization accuracy (centroid distance) and lesion detectability on the LiTS dataset (avg. min. detectable volume).}
\label{tab:lits_spatial}
\footnotesize
\begin{tabular}{l cccc}
\toprule
 & \multicolumn{2}{c}{\textbf{Liver}} & \multicolumn{2}{c}{\textbf{Tumor}} \\
\cmidrule(r){2-3} \cmidrule(l){4-5}
Method & CD (mm) $\downarrow$ & Avg. $V_{\text{min}}$ ($\text{mm}^3$) $\downarrow$ & CD (mm) $\downarrow$ & Avg. $V_{\text{min}}$ ($\text{mm}^3$) $\downarrow$ \\
\midrule
SegFormer       & 5.32$\pm$7.64 & 715,136.15 & 15.67$\pm$13.29 & 1,645.88 \\
Dino U-Net      & 6.01$\pm$15.02 & 715,136.22 & 20.43$\pm$18.56 & \textbf{1,596.45} \\
nnU-Net         & 4.04$\pm$5.71 & 618,018.85 & 11.26$\pm$11.45 & 1,625.59 \\
MedDINOv3       & \textbf{1.07}$\pm$0.28 & \textbf{605,396.66} & 7.98$\pm$4.47 & 4,799.19 \\
\textbf{DynoDINO (ours)} & 1.20$\pm$0.29 & \textbf{605,396.66} & \textbf{7.56}$\pm$5.47 & 4,826.23 \\
\bottomrule
\end{tabular}
\end{table*}


\begin{table*}[t]
\centering
\caption{Quantitative segmentation performance on the PLC-CECT dataset. 
}
\label{tab:plc_cect_results}
\footnotesize
\setlength{\tabcolsep}{5pt}
\begin{tabular}{l cccccc c}
\toprule
 & \multicolumn{2}{c}{\textbf{Liver}} & \multicolumn{2}{c}{\textbf{Tumor}} & \multicolumn{3}{c}{\textbf{Global Performance}} \\
\cmidrule(r){2-3} \cmidrule(lr){4-5} \cmidrule(l){6-8}
Method & DSC (\%) $\uparrow$ & Sens. (\%) $\uparrow$ & DSC (\%) $\uparrow$ & Sens. (\%) $\uparrow$ & DSC (\%) $\uparrow$ & NSD (\%) $\uparrow$ & HD95 (mm) $\downarrow$ \\
\midrule
\multicolumn{8}{@{}l}{\textit{Single-phase Input}} \\
SegFormer       & 87.47$\pm$2.74 & 91.57$\pm$2.30 & 58.44$\pm$8.59 & 53.48$\pm$9.11 & 72.96 & 38.23 & 32.72 \\
Dino U-Net      & 87.52$\pm$3.23 & 93.14$\pm$2.57 & 60.75$\pm$7.93 & 56.11$\pm$9.22 & 74.14 & 39.50 & 32.41 \\
nnU-Net         & \textbf{88.71}$\pm$3.15 & 92.77$\pm$1.92 & \textbf{64.64}$\pm$8.09 & \textbf{62.18}$\pm$8.94 & \textbf{76.68} & 44.52 & 29.71 \\
MedDINOv3       & 81.77$\pm$4.31 & 84.64$\pm$3.74 & 52.04$\pm$9.39 & 44.72$\pm$9.50 & 66.91 & 21.95 & 33.41 \\
\textbf{DynoDINO (ours)} & 88.38$\pm$3.16 & \textbf{94.23}$\pm$1.77 & 56.41$\pm$9.27 & 47.23$\pm$9.01 & 72.40 & \textbf{44.67} & \textbf{26.81} \\
\midrule
\multicolumn{8}{@{}l}{\textit{Multi-phase Input}} \\
nnU-Net         & 88.92$\pm$2.92 & 92.73$\pm$2.51 & \textbf{69.76}$\pm$6.88 & \textbf{66.23}$\pm$8.15 & \textbf{79.34} & 44.60 & 28.72 \\
\textbf{DynoDINO (ours)} & \textbf{89.77}$\pm$2.79 & \textbf{93.40}$\pm$2.67 & 66.23$\pm$8.42 & 61.67$\pm$9.42 & 78.00 & \textbf{47.10} & \textbf{24.35} \\
\bottomrule
\end{tabular}
\end{table*}

\begin{table*}[t]
\centering
\caption{Spatial localization accuracy and lesion detectability on the PLC-CECT dataset. 
}
\label{tab:plc_cect_spatial}
\footnotesize
\begin{tabular}{l cccc}
\toprule
 & \multicolumn{2}{c}{\textbf{Liver}} & \multicolumn{2}{c}{\textbf{Tumor}} \\
\cmidrule(r){2-3} \cmidrule(l){4-5}
Method & CD (mm) $\downarrow$ & Avg. $V_{\text{min}}$ ($\text{mm}^3$) $\downarrow$ & CD (mm) $\downarrow$ & Avg. $V_{\text{min}}$ ($\text{mm}^3$) $\downarrow$ \\
\midrule
\multicolumn{5}{l}{\textit{Single-phase Input}} \\
SegFormer       & \textbf{10.14}$\pm$3.52 & 166,908.89 & 15.78$\pm$6.10 & 481,747.28 \\
Dino U-Net      & 11.67$\pm$4.71 & 258,009.22 & 15.27$\pm$5.26 & \textbf{463,409.57} \\
nnU-Net         & 10.24$\pm$5.67 & 97,864.70 & \textbf{10.60}$\pm$3.18 & 466,601.66 \\
MedDINOv3       & 13.93$\pm$6.27 & 104,759.92 & 24.98$\pm$11.27 & 562,741.91 \\
\textbf{DynoDINO (ours)} & 10.17$\pm$4.13 & \textbf{90,145.65} & 11.20$\pm$2.45 & 558,828.88 \\
\midrule
\multicolumn{5}{l}{\textit{Multi-phase Input}} \\
nnU-Net         & 9.30$\pm$3.87 & 26,295.70 & 13.46$\pm$9.14 & \textbf{470,586.67} \\
\textbf{DynoDINO (ours)} & \textbf{8.47}$\pm$3.30 & \textbf{21,871.95} & \textbf{12.05}$\pm$7.68 & 477,496.58 \\
\bottomrule
\end{tabular}
\end{table*}


\begin{table*}[t]
\centering
\caption{Quantitative segmentation performance on the WAW-TACE dataset. 
}
\label{tab:waw_tace_results}
\footnotesize
\setlength{\tabcolsep}{5pt}
\begin{tabular}{l cccccc c}
\toprule
 & \multicolumn{2}{c}{\textbf{Liver}} & \multicolumn{2}{c}{\textbf{Tumor}} & \multicolumn{3}{c}{\textbf{Global Performance}} \\
\cmidrule(r){2-3} \cmidrule(lr){4-5} \cmidrule(l){6-8}
Method & DSC (\%) $\uparrow$ & Sens. (\%) $\uparrow$ & DSC (\%) $\uparrow$ & Sens. (\%) $\uparrow$ & DSC (\%) $\uparrow$ & NSD (\%) $\uparrow$ & HD95 (mm) $\downarrow$ \\
\midrule
\multicolumn{8}{@{}l}{\textit{Single-phase Input}} \\
SegFormer       & 94.13$\pm$1.97 & 95.98$\pm$0.82 & 45.20$\pm$16.50 & 36.47$\pm$15.99 & 69.67 & 44.39 & 31.96 \\
Dino U-Net      & 93.89$\pm$2.02 & 94.84$\pm$1.48 & 53.46$\pm$15.73 & 49.29$\pm$17.78 & 73.68 & 42.25 & 44.37 \\
nnU-Net         & \textbf{96.01}$\pm$1.55 & 96.91$\pm$0.64 & \textbf{62.06}$\pm$15.31 & \textbf{61.19}$\pm$17.89 & \textbf{79.04} & \textbf{54.19} & 21.31 \\
MedDINOv3       & 92.90$\pm$1.87 & 93.25$\pm$2.30 & 21.12$\pm$15.48 & 18.06$\pm$16.82 & 57.01 & 31.40 & 22.35 \\
\textbf{DynoDINO (ours)} & 95.85$\pm$1.70 & \textbf{97.80}$\pm$0.45 & 43.58$\pm$20.12 & 38.75$\pm$20.56 & 69.72 & 52.10 & \textbf{19.20} \\
\midrule
\multicolumn{8}{@{}l}{\textit{Multi-phase Input}} \\
nnU-Net         & 95.94$\pm$1.30 & 96.99$\pm$0.61 & \textbf{59.56}$\pm$15.34 & \textbf{56.05}$\pm$16.98 & \textbf{77.75} & 53.11 & 20.18 \\
\textbf{DynoDINO (ours)} & \textbf{96.14}$\pm$1.34 & \textbf{97.46}$\pm$0.54 & 44.28$\pm$19.71 & 39.04$\pm$20.50 & 70.21 & \textbf{54.69} & \textbf{18.54} \\
\bottomrule
\end{tabular}
\end{table*}

\begin{table*}[t]
\centering
\caption{Spatial localization accuracy and lesion detectability on the WAW-TACE dataset. 
}
\label{tab:waw_tace_spatial}
\footnotesize
\begin{tabular}{l cccc}
\toprule
 & \multicolumn{2}{c}{\textbf{Liver}} & \multicolumn{2}{c}{\textbf{Tumor}} \\
\cmidrule(r){2-3} \cmidrule(l){4-5}
Method & CD (mm) $\downarrow$ & Avg. $V_{\text{min}}$ ($\text{mm}^3$) $\downarrow$ & CD (mm) $\downarrow$ & Avg. $V_{\text{min}}$ ($\text{mm}^3$) $\downarrow$ \\
\midrule
\multicolumn{5}{l}{\textit{Single-phase Input}} \\
SegFormer       & 6.00$\pm$6.89 & 1,194,212.54 & 28.34$\pm$30.91 & 31,507.25 \\
Dino U-Net      & 5.68$\pm$3.22 & 1,396,202.08 & 30.95$\pm$19.73 & \textbf{25,087.62} \\
nnU-Net         & \textbf{1.71}$\pm$1.2 & 1,302,667.15 & \textbf{15.15}$\pm$7.86 & 26,025.85 \\
MedDINOv3       & 3.82$\pm$1.74 & \textbf{1,194,212.54} & 43.39$\pm$23.46 & 32,598.90 \\
\textbf{DynoDINO (ours)} & 1.86$\pm$1.49 & \textbf{1,194,212.54} & 27.94$\pm$16.14 & 30,641.27 \\
\midrule
\multicolumn{5}{l}{\textit{Multi-phase Input}} \\
nnU-Net         & 2.34$\pm$1.62 & \textbf{1,194,182.00}  & \textbf{16.85}$\pm$11.31 & \textbf{25,942.31} \\
\textbf{DynoDINO (ours)} & \textbf{1.61}$\pm$1.21 & 1,194,212.54 & 22.36$\pm$13.82 & 28,040.67 \\
\bottomrule
\end{tabular}
\end{table*}

\subsection{Qualitative Analysis}
\label{sec:qual-analysis}

\begin{figure*}[t]
\centering
\includegraphics[width=1.0\linewidth]{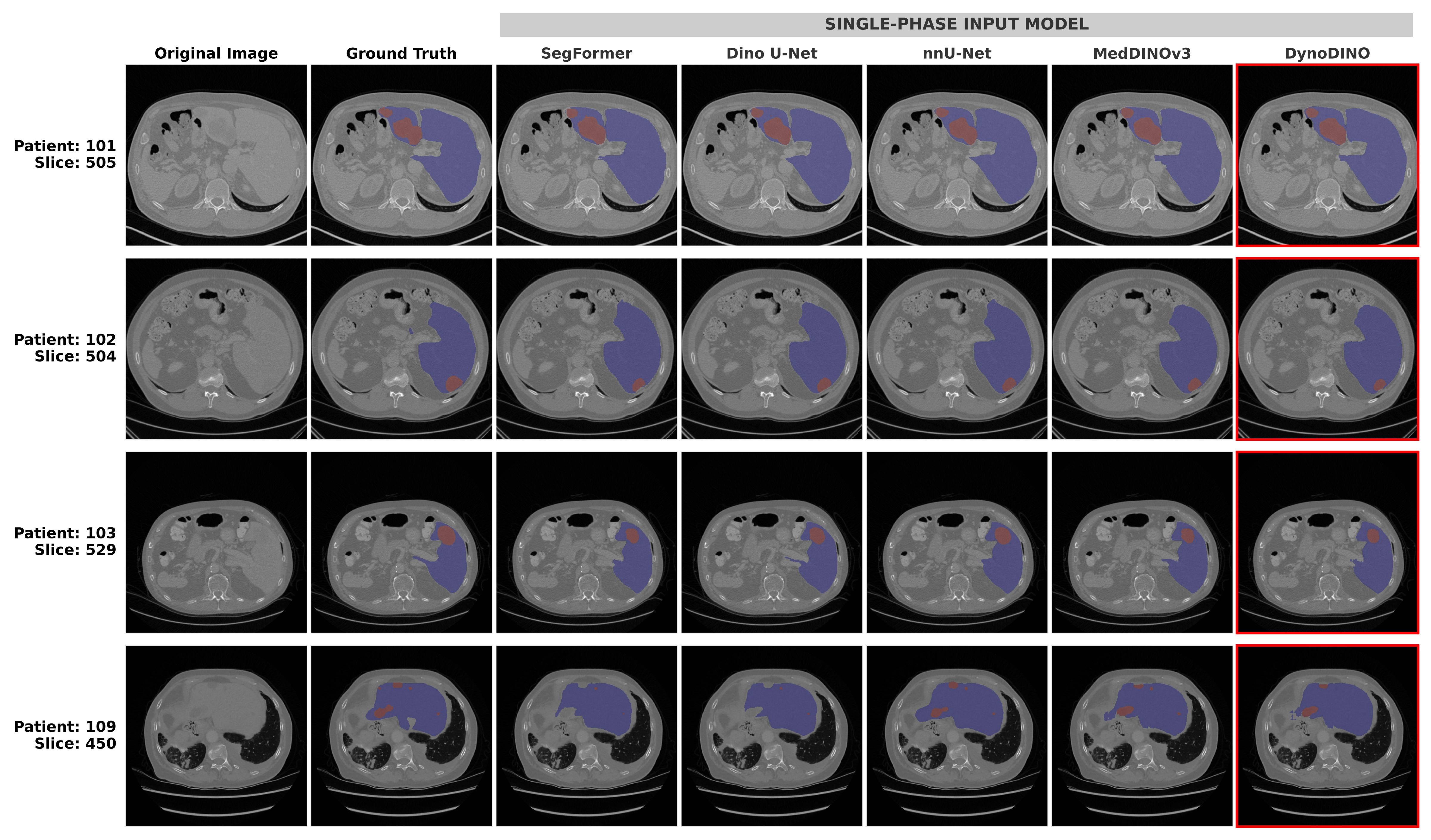}
\caption{Qualitative comparison of segmentation results on the LiTS dataset. Results are shown in red.}
\label{fig:lits_viz}
\end{figure*}

\begin{figure*}[t]
\centering
\includegraphics[width=1.0\linewidth]{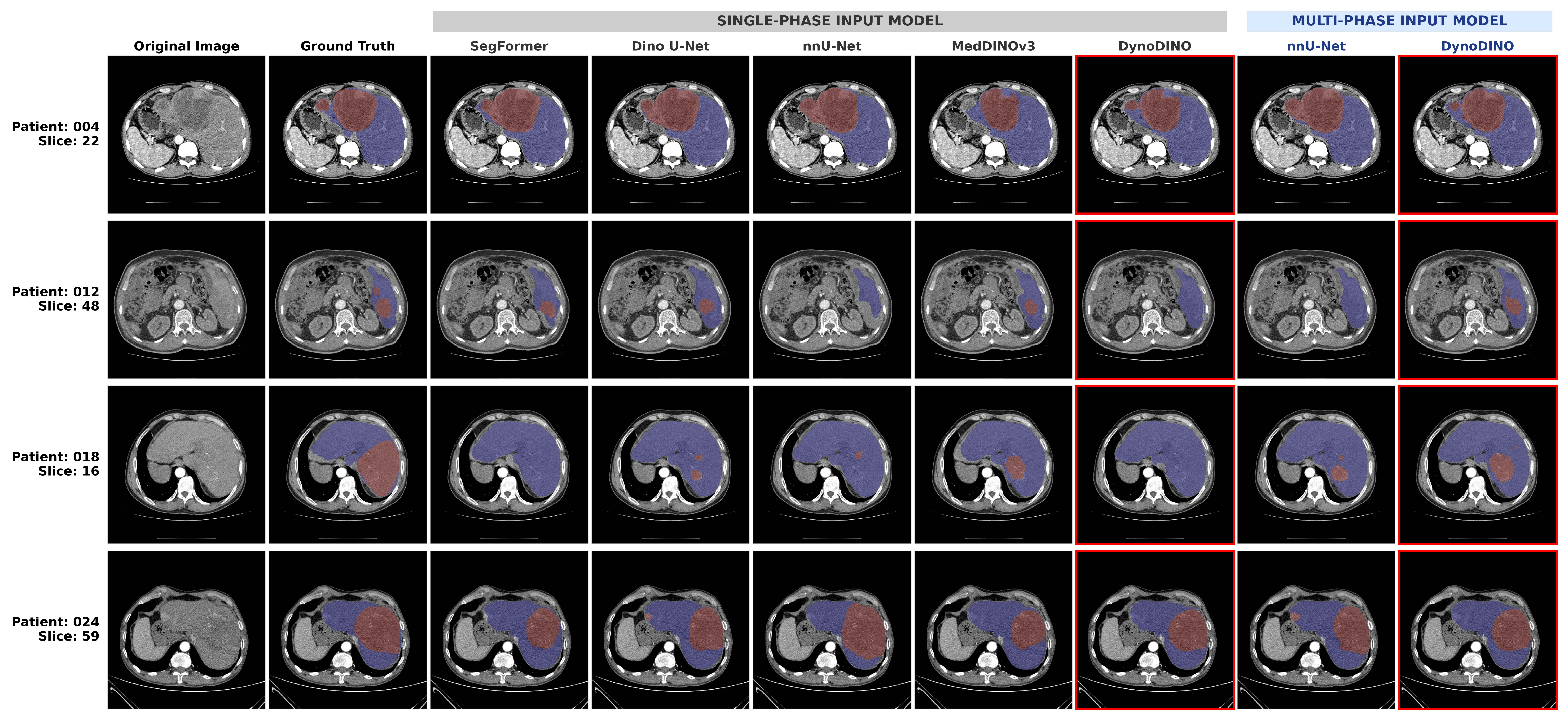}
\caption{Qualitative comparison of segmentation results on the PLC-CECT dataset.}
\label{fig:plc_viz}
\end{figure*}

\begin{figure*}[t]
\centering
\includegraphics[width=1.0\linewidth]{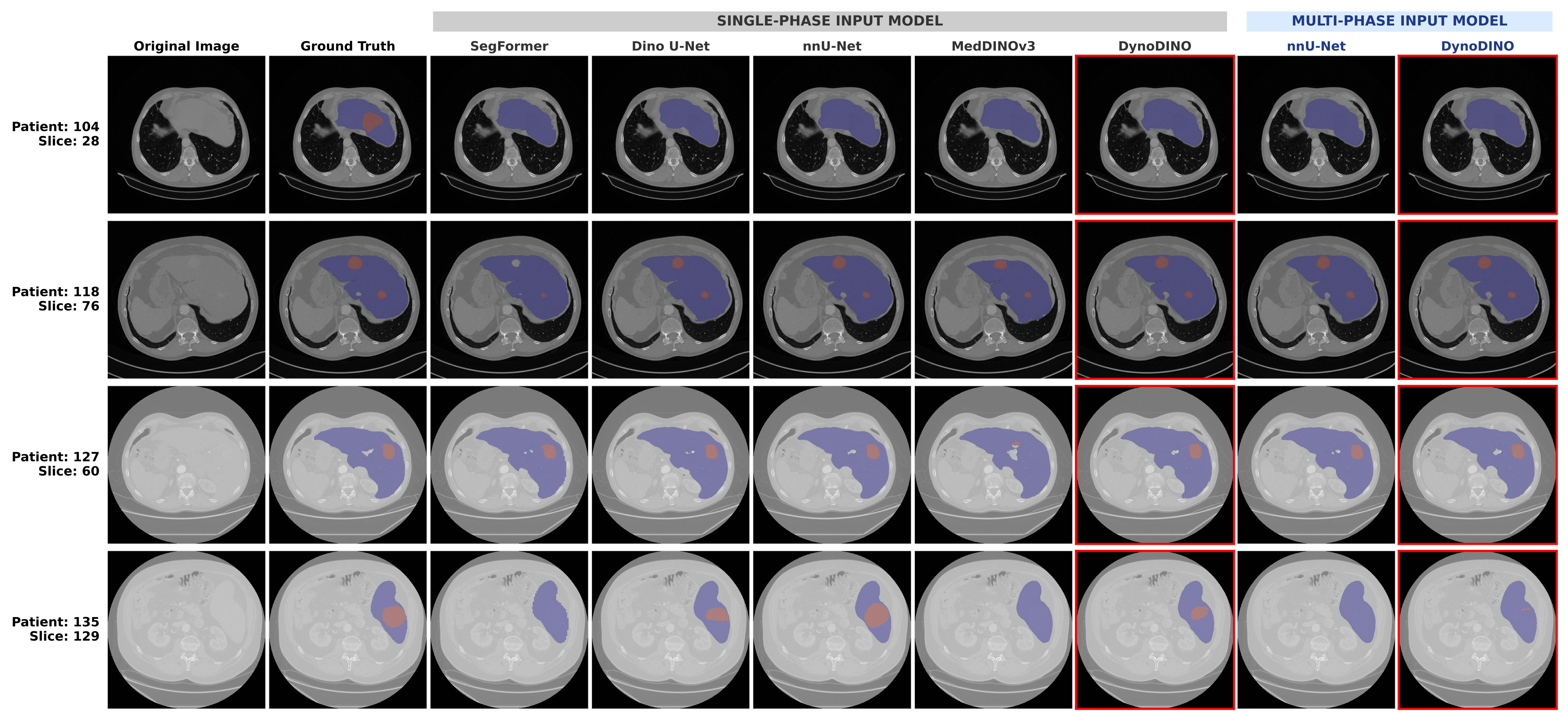}
\caption{Qualitative comparison of segmentation results on the WAW-TACE dataset.}
\label{fig:waw_viz}
\end{figure*}

We present visual comparative results across the LiTS (Fig.~\ref{fig:lits_viz}), PLC-CECT (Fig.~\ref{fig:plc_viz}), and WAW-TACE (Fig.~\ref{fig:waw_viz}) cohorts to evaluate the clinical utility of our framework.

\textbf{Model-Specific Observations.}
Across all evaluated datasets, distinct architectural behaviors emerge under visual inspection. Standard nnU-Net exhibits a persistent \textbf{over-smoothing bias}; while it effectively covers core regional masks, it frequently sacrifices fine-grained, non-convex morphological details. Conversely, SegFormer consistently suffers from \textbf{boundary coarseness} and localized under-segmentation. The self-supervised baseline, MedDINOv3, although structurally robust, frequently struggles with \textbf{sub-optimal boundary delineation} in transitional tumor zones.

\textbf{Dataset-Specific Observations.}
In the public LiTS dataset (Fig.~\ref{fig:lits_viz}), Patient 101 highlights these architectural trade-offs: SegFormer and MedDINOv3 yield highly imprecise tumor margins, while Dino U-Net introduces an isolated, false-positive micro-tumor artifact directly superior to the large primary tumor. For the non-convex tumor margins observed in Patients 101–103, nnU-Net's smoothing bias leads to structural erosion, whereas DynoDINO maintains superior morphological fidelity, adhering tightly to complex liver boundaries.
However, the multi-focal scenario of Patient 109 exposes a clear performance disparity and boundary limitation under single-phase constraints. Notably, all evaluated models completely fail to detect the subtle micro-lesion located at the top-leftmost margin of the liver. In this challenging case, standard nnU-Net delivers the superior baseline performance, capturing the multi-focal distribution with high regional fidelity due to its dense local convolutions. MedDINOv3 follows as the second-best architecture, retaining the capacity to detect the primary tumor alongside the smaller peripheral clusters. In contrast, DynoDINO only resolves the large primary tumor mass while entirely omitting the smaller micro-lesions. This qualitative outcome visually corroborates our quantitative findings in Table~\ref{tab:lits_spatial}. It demonstrates that under strict single-phase constraints—where temporal cues are absent—DynoDINO's structural regularizers tend to heavily prioritize macroscopic boundary cohesion, inadvertently dismissing isolated micro-lesions as background noise. This limitation underscores that multi-phase temporal synergy is fundamentally required to fully unlock the framework's localized detection sensitivity.

In the clinical PLC-CECT cohort (Fig.~\ref{fig:plc_viz}), the integration of multi-phase temporal synergy overcomes these single-phase limitations. Patient 004 showcases the topographical precision unlocked by our framework; only DynoDINO successfully differentiates two independent, adjacent lesions, whereas all baseline models either merge them into a single mass or omit them entirely. A more profound clinical advantage is observed in Patient 012, where the gold-standard nnU-Net completely fails to segment any tumor structures. Among all tested networks, only the multi-phase configuration of DynoDINO successfully and correctly localizes both independent lesions. Furthermore, for Patient 018, DynoDINO's segmentation area is visually the closest to the expert Ground Truth. In Patient 024, SegFormer exhibits severe localized under-segmentation along the tumor periphery, and both Dino U-Net and nnU-Net generate substantial false-positive regions in healthy tissue, while DynoDINO remains tightly constrained to the true pathological boundaries.

The post-treatment WAW-TACE cohort confirms the resilience of DynoDINO under extreme imaging noise, severe artifacts, and lipiodol depositions (Fig.~\ref{fig:waw_viz}). In the highly challenging case of Patient 135, a compelling architectural anomaly is observed: while the single-phase nnU-Net yields acceptable baseline segments, its multi-phase counterpart completely fails to detect or segment the tumor due to the introduced step-artifacts from brute-force padding. In contrast, DynoDINO successfully identifies the lesion under both single- and multi-phase configurations. For Patient 104, although the severe post-treatment artifacts cause all models to completely omit the true tumor regions, DynoDINO clearly delineates the most accurate and anatomically precise liver boundary among all baselines. Finally, across Patients 118 and 127, DynoDINO achieves high localization precision and sharp edge delineation without generating the over-smoothing artifacts characteristic of nnU-Net.

\subsection{Ablation Study}


We investigate the following core architectural choices.
\begin{enumerate}
    \item 
    \textbf{Inter-Phase Fusion Method}: Evaluating the performance of \textbf{MA} against vanilla Cross Attention when synthesizing information between the reference (LP) and target (EP/MP) feature streams.
    \item 
    \textbf{Kinetic Modeling Strategy}: Comparing the effectiveness of the \textbf{Subtraction-based dynamic modeling} against standard Cross Attention for capturing dynamic contrast signatures.
    \item 
    \textbf{Adaptive Gating Mechanism}: Verifying whether the integration of a Sigmoid-based gating function prior to the residual skip connection effectively filters noise.
\end{enumerate}

\subsubsection{Analysis of Ablated Components}

The evaluated variants and their characteristics are summarized in Table~\ref{tab:ablation}.
Our key observations are as follows.

\textbf{Kinetic Modeling via Subtraction.} 
Comparing Variants A and B, the kinetic modeling achieves higher region accuracy (DSC 77.17\%), effectively mimicking clinical subtraction logic while remaining efficient.

\textbf{Inter-Phase Mix-Attention.} 
MA (Variants C and D) reduces fusion parameters by 7.08M, while maintaining high-resolution feature maps during iterative refinement.

\textbf{Adaptive Gating.} 
Transitioning from Variant D to Proposed, NSD is improved from 46.00\% to 47.10\%, validating its role in handling misalignment.

\begin{table*}[t]
\centering
\footnotesize
\setlength{\tabcolsep}{3.5pt}
\caption{Ablation study on phase fusion methods, difference extraction, and gating mechanisms (PLC-CECT).}
\label{tab:ablation}
\begin{tabular}{l cccc ccc}
\toprule
\textbf{Model} & \textbf{Inter-Phase Fusion} & \textbf{Kinetic Modeling} & \textbf{Gating} & Params. $\downarrow$ & DSC (\%) $\uparrow$ & NSD (\%) $\uparrow$ & HD95 (mm) $\downarrow$ \\
\midrule
Variant A & Cross-attn. & Cross-attn. & \ding{55} & 170,929,155 & 76.82 & \textbf{47.25} & \textbf{23.90} \\
Variant B & Cross-attn. & Subtraction & \ding{55} & 168,569,091 & \textbf{77.17} & 46.01 & 24.79 \\
Variant C & Mix-attn.   & Cross-attn. & \ding{55} & 163,851,267 & 76.92 & 46.99 & 24.03 \\
Variant D & Mix-attn.   & Subtraction & \ding{55} & \textbf{161,491,203} & 76.09 & 46.00 & 24.74 \\
\midrule
\textbf{DynoDINO (ours)} & Mix-attn. & Subtraction & \ding{51} & 163,852,803 & 76.01 & 47.10 & 24.35 \\
\bottomrule
\end{tabular}
\end{table*}

\subsubsection{Fusion Strategy: Difference vs. Concatenation}

DynoDINO employs a difference-based residual learning strategy. We compare it against two other fusion paradigms: 
(i) \textit{Naive Concatenation}, where feature maps from all phases are simply stacked, and 
(ii) \textit{Concatenation + Explicit Differences}, which augments the concatenated features with pairwise subtracted feature maps.

While Naive Concatenation achieves competitive DSC scores (76.10\% in PLC-CECT), it struggles with boundary localization, resulting in the worst HD95 across both datasets (Table~\ref{tab:fusion_strategy_comp}). 
Adding explicit differences provides marginal improvements in WAW-TACE, but does not significantly enhance boundary fidelity. 
In contrast, DynoDINO consistently achieves the best performance in surface-based metrics (HD95=24.35 mm in PLC-CECT and 17.79 mm in WAW-TACE). These results indicate that our difference-based residual fusion strategy more effectively isolates and translates temporal contrast kinetics into explicit structural constraints than generic feature stacking. By directly computing inter-phase feature subtractions, the network is regularized to capture subtle hemodynamic variations rather than relying on unconstrained channel-wise concatenations, ultimately  leading to segmentations that are more anatomically precise and clinically reliable.

\begin{table}[t]
\centering
\footnotesize
\caption{Ablation of fusion strategies (PLC-CECT and WAW-TACE).}
\label{tab:fusion_strategy_comp}
\begin{tabular}{l ccc ccc}
\toprule
& \multicolumn{3}{c}{\textbf{PLC-CECT}} & \multicolumn{3}{c}{\textbf{WAW-TACE}} \\
\cmidrule(lr){2-4} \cmidrule(lr){5-7}
Fusion Strategy & DSC (\%) $\uparrow$ & NSD (\%) $\uparrow$ & HD95 (mm) $\downarrow$ & DSC (\%) $\uparrow$ & NSD (\%) $\uparrow$ & HD95 (mm) $\downarrow$ \\
\midrule
Concatenation            & 76.10 & 45.02 & 26.05 & 76.93 & 52.17 & 20.58 \\
Explicit Diff. + Concat  & \textbf{76.29} & 44.84 & 26.19 & 76.81 & 52.84 & 20.12 \\
\textbf{Fusion Model (Ours)} & 76.01 & \textbf{47.10} & \textbf{24.35} & \textbf{77.33} & \textbf{54.15} & \textbf{17.79} \\
\bottomrule
\end{tabular}
\end{table}

\subsubsection{Sensitivity to Missing Phase (Phase Dropout)}

To validate the model's reliance on temporal information, we conduct a ``Phase Dropout'' analysis during the inference stage (Table~\ref{tab:phase_dropout}). 

In the PLC-CECT cohort, performance scales linearly with the number of available phases, achieving the most optimal boundary fidelity (HD95: 24.35 mm) when complete triphasic data is provided. Conversely, a catastrophic performance collapse is observed when the EP is omitted, resulting in a tumor DSC of merely 0.59\%. This severe degradation firmly confirms that DynoDINO strictly utilizes the EP as its foundational reference frame for difference-based residual learning. Removing this crucial spatial anchor during inference violates the rigid structural dependencies established during training on complete triphasic distributions (Table~\ref{tab:dataset_dist}).

In contrast, evaluations on the WAW-TACE dataset demonstrate a more resilient and nuanced integration of temporal features. While the EP-only baseline already yields a robust performance (DSC: 76.52\%), introducing the MP or LP sequences provides incremental gains in surface alignment (HD95 improving from 18.82 mm to 18.53 mm). The optimal topological precision is ultimately unlocked under the full triphasic configuration (DSC: 77.33\%, HD95: 17.79 mm).

This distinct behaviors across cohorts indicate that under severe clinical artifacts and inconsistent scanning intervals, the Adaptive Gating mechanism robustly prevents the network from collapsing. Instead of failing catastrophically in the absence of complete sequences, DynoDINO dynamically routes and aggregates whatever partial temporal cues are available during inference, demonstrating superior clinical viability and adaptive robustness for real-world deployment.

\begin{table}[t]
\centering
\footnotesize
\caption{Sensitivity analysis to phase dropout during inference.}
\label{tab:phase_dropout}
\begin{tabular}{l ccc ccc}
\toprule
& \multicolumn{3}{c}{\textbf{PLC-CECT}} & \multicolumn{3}{c}{\textbf{WAW-TACE}} \\
\cmidrule(lr){2-4} \cmidrule(lr){5-7}
Combination & DSC (\%) $\uparrow$ & NSD (\%) $\uparrow$ & HD95 (mm) $\downarrow$ & DSC (\%) $\uparrow$ & NSD (\%) $\uparrow$ & HD95 (mm) $\downarrow$ \\
\midrule
EP only          & 62.48 & 39.42 & 32.16 & 76.52 & 54.21 & 18.82 \\
EP + MP          & 69.11 & 41.18 & 29.18 & 76.78 & 54.11 & 18.59 \\
EP + LP          & 73.07 & 45.90 & 25.33 & 76.86 & 53.80 & 18.53 \\
MP + LP          & 0.59  & 0.19  & 79.58 & 58.22 & 12.20 & 27.37 \\
\textbf{EP+MP+LP} & \textbf{76.01} & \textbf{47.10} & \textbf{24.35} & \textbf{77.33} & \textbf{54.15} & \textbf{17.79} \\
\bottomrule
\end{tabular}
\end{table}

\subsubsection{Robustness to Misalignment and Spatial Noise}
A key clinical requirement is robustness against respiratory-induced motion and inter-phase misalignment. We perform an ablation study of the FFT-based pre-alignment and the Adaptive Gating Mechanism from four perspectives.

\paragraph{Justification of ZNCC for Pre-alignment.}
To rigorously justify the selection of ZNCC for our FFT-based pre-alignment module, we established a controlled axial displacement simulation framework using a subset of the PLC-CECT cohort. Specifically, 30 patients were randomly selected, and 30 independent axial slices were sampled per patient (yielding 900 evaluation instances). With the EP slice held fixed as the spatial anchor, artificial random axial translations ranging within $\pm10$ slices were independently introduced to the corresponding MP and LP sequences. 

We systematically evaluated the capability of six prominent similarity metrics—Cosine Similarity, Mean Absolute Difference (MAD), Mean Squared Difference (MSD), Mutual Information (MI), Structural Similarity Index (SSIM), and ZNCC—to recover these induced spatial offsets (Table~\ref{tab:metric_alignment_slices}). Standard structural and intensity-based metrics struggled to handle the complex contrast dynamics. Notably, SSIM exhibited the lowest tracking accuracy (58.77\%), while MI suffered from a severe Mean Absolute Error (MAE) of 0.916 $\pm$ 2.14 slices. This degradation is primarily attributed to the radical, non-linear intensity variations and localized contrast inversions inherent to multi-phase contrast-medium kinetics, which introduce misleading local optima for statistical and structure-matching paradigms. 

In contrast, ZNCC achieved the highest localization alignment accuracy (65.25\%) and minimized the alignment error to a sub-slice level of 0.718 $\pm$ 1.78 slices, marginally outperforming Cosine Similarity (0.720 $\pm$ 1.78 slices). This sub-slice resolution demonstrates that by intrinsically normalizing local mean intensities and variances, ZNCC effectively filters out global contrast shifts caused by hemodynamics, establishing itself as the most resilient mathematical regularizer for mitigating inter-phase respiratory motion in our framework.

\begin{table}[t]
\centering
\footnotesize
\caption{Quantitative comparison of image similarity metrics under induced multi-phase axial misalignments ($\pm10$ slices search space).}
\label{tab:metric_alignment_slices}
\begin{tabular}{l cc}
\toprule
\textbf{Similarity Metric} & MAE (Slices) $\downarrow$ & Tracking Accuracy (\%) $\uparrow$ \\
\midrule
Cosine               & 0.720 $\pm$ 1.78 & 64.69 \\
MAD                  & 0.791 $\pm$ 1.95 & 62.65 \\
MSD                  & 0.767 $\pm$ 1.84 & 62.10 \\
MI                   & 0.916 $\pm$ 2.14 & 61.36 \\
SSIM                 & 0.856 $\pm$ 1.86 & 58.77 \\
\textbf{ZNCC (Ours)} & \textbf{0.718} $\pm$ 1.78 & \textbf{65.25} \\
\bottomrule
\end{tabular}
\end{table}

\paragraph{Impact of FFT-ZNCC Alignment.} 
We evaluated the necessity of synchronization by ablating the FFT-ZNCC preprocessing step (Table~\ref{tab:fft_ablation}). In PLC-CECT, the model without FFT-ZNCC performs comparably to the pre-aligned version, suggesting that for standard diagnostic scans, our Adaptive Gating provides sufficient spatial tolerance. However, the WAW-TACE cohort exhibited significant degradation: DSC dropped from 77.33\% to 75.28\%, and HD95 deteriorated to 19.74 mm. These findings underscore that FFT-ZNCC is critical for ``messy'' clinical data with inconsistent breath-holds.

\begin{table}[t]
\centering
\footnotesize
\caption{Ablation of FFT-ZNCC pre-alignment.}
\label{tab:fft_ablation}
\begin{tabular}{l ccc ccc}
\toprule
& \multicolumn{3}{c}{\textbf{PLC-CECT}} & \multicolumn{3}{c}{\textbf{WAW-TACE}} \\
\cmidrule(lr){2-4} \cmidrule(lr){5-7}
Alignment & DSC (\%) $\uparrow$ & NSD (\%) $\uparrow$ & HD95 (mm) $\downarrow$ & DSC (\%) $\uparrow$ & NSD (\%) $\uparrow$ & HD95 (mm) $\downarrow$ \\
\midrule
w/o FFT-ZNCC         & \textbf{76.51} & 46.28 & \textbf{24.35} & 75.28 & \textbf{55.50} & 19.74 \\
\textbf{w/ FFT-ZNCC (Ours)} & 76.01 & \textbf{47.10} & \textbf{24.35} & \textbf{77.33} & 54.15 & \textbf{17.79} \\
\bottomrule
\end{tabular}
\end{table}

\paragraph{Robustness to In-plane Shift.} 
To simulate extreme registration errors, we applied artificial 2D translations to the MP and LP sequences during inference (Table~\ref{tab:robustness}). The results reveal a critical dependency on the Adaptive Gating module. 
In PLC-CECT, while nnU-Net maintains a slightly higher DSC, it exhibits significantly poorer boundary precision (HD95: 29.35 mm vs. 24.69 mm). More importantly, in the WAW-TACE cohort, the model without Adaptive Gating suffered a \textbf{catastrophic failure} (DSC: 0.00\%), whereas the full DynoDINO model remained robust (DSC: 76.81\%, HD95: 18.61 mm). This disparity highlights a fundamental risk: without a gating mechanism to filter out spatial noise, misaligned features are amplified through the subtraction logic, leading to divergent latent representations.

\begin{table}[t]
\centering
\footnotesize
\caption{Robustness analysis against in-plane spatial shifts.}
\label{tab:robustness}
\begin{tabular}{l ccc ccc}
\toprule
& \multicolumn{3}{c}{\textbf{PLC-CECT}} & \multicolumn{3}{c}{\textbf{WAW-TACE}} \\
\cmidrule(lr){2-4} \cmidrule(lr){5-7}
Configuration & DSC (\%) $\uparrow$ & NSD (\%) $\uparrow$ & HD95 (mm) $\downarrow$ & DSC (\%) $\uparrow$ & NSD (\%) $\uparrow$ & HD95 (mm) $\downarrow$ \\
\midrule
nnU-Net                  & \textbf{76.95} & 44.48 & 29.35 & \textbf{79.11} & 53.16 & 20.42 \\
DynoDINO w/o Gating      & 76.46 & 46.36 & 24.89 & 0.00 & 0.00 & N/A \\
\textbf{DynoDINO (Ours)} & 76.32 & \textbf{46.60} & \textbf{24.69} & 76.81 & \textbf{54.19} & \textbf{18.61} \\
\bottomrule
\end{tabular}
\end{table}

\begin{figure*}[t]
\centering
\begin{subfigure}{0.48\textwidth}
    \centering
    \includegraphics[width=\linewidth]{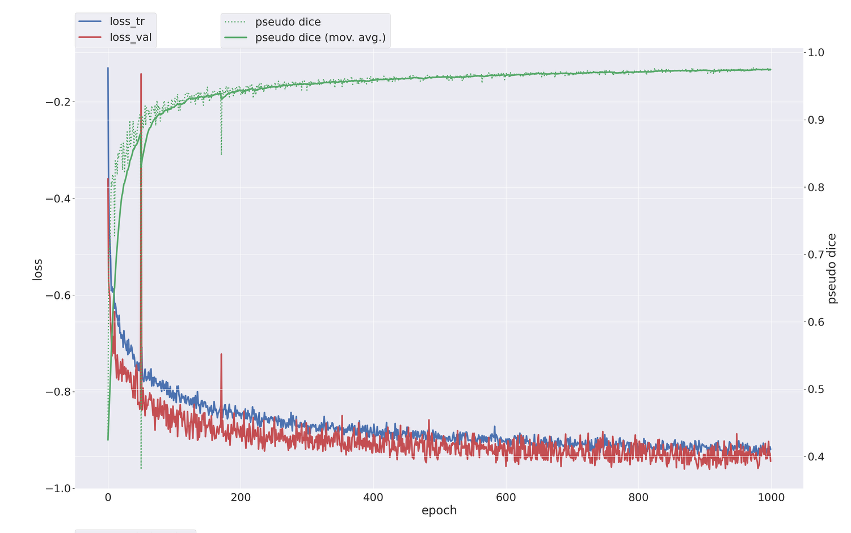}
    \caption{No Shift + Gating (Standard)}
\end{subfigure}
\hfill
\begin{subfigure}{0.48\textwidth}
    \centering
    \includegraphics[width=\linewidth]{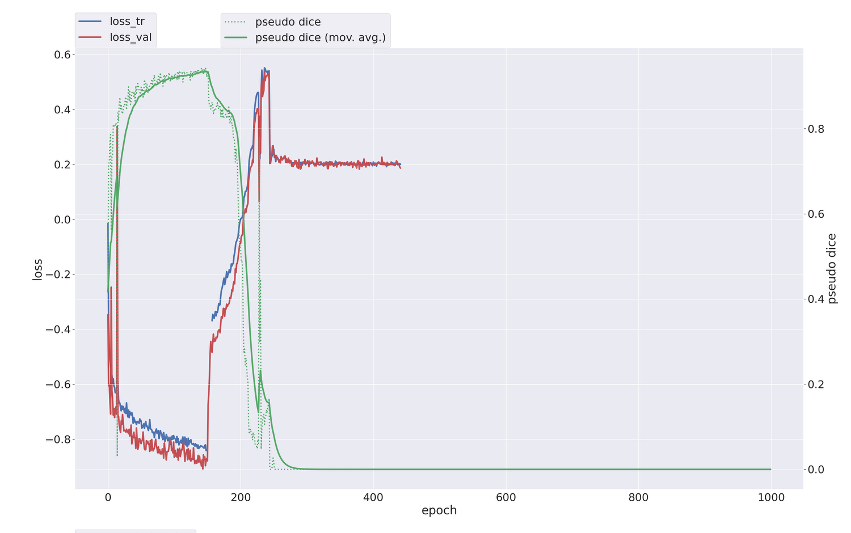}
    \caption{No Shift, No Gating (Ablated)}
\end{subfigure}

\vspace{1em}

\begin{subfigure}{0.48\textwidth}
    \centering
    \includegraphics[width=\linewidth]{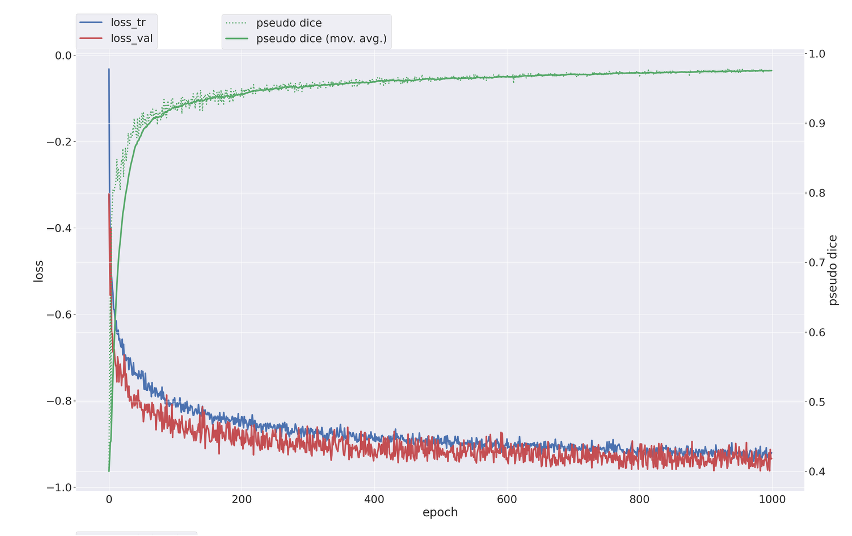}
    \caption{Shifted + Gating (Proposed)}
\end{subfigure}
\hfill
\begin{subfigure}{0.48\textwidth}
    \centering
    \includegraphics[width=\linewidth]{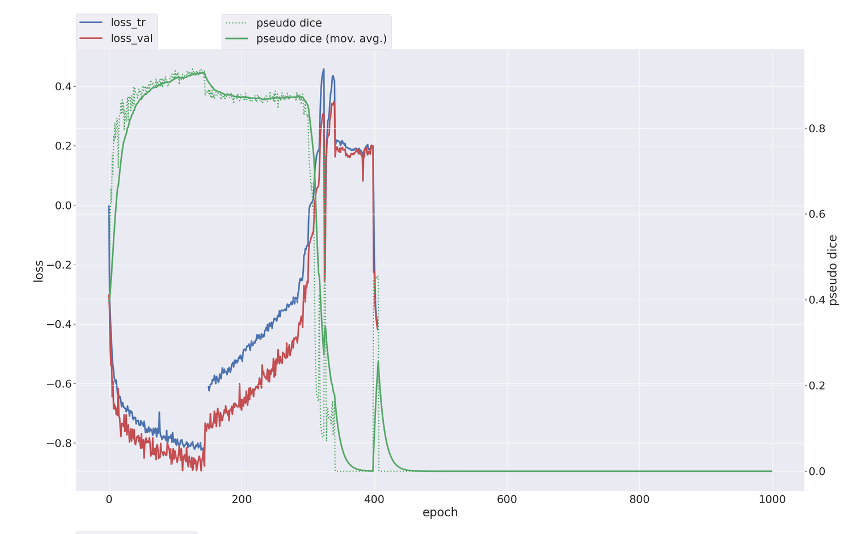}
    \caption{Shifted, No Gating (Collapse)}
\end{subfigure}
\caption{Analysis of training dynamics and convergence stability on the WAW-TACE dataset. Blue and red curves represent training and validation loss, respectively. Solid green curves denote the Pseudo Dice score, with dashed lines showing the moving average. 
(a) standard DynoDINO configuration with gating and no spatial shift, (b) ablated model without gating and no spatial shift, (c) proposed DynoDINO configuration with gating under simulated in-plane spatial shift, and (d) ablated model without gating under simulated in-plane spatial shift.}
\label{fig:loss_dynamics}
\end{figure*}

\paragraph{Learning Stability and Convergence Analysis.}
To further investigate the optimization failure observed in Table~\ref{tab:robustness}, we analyzed the training dynamics on the WAW-TACE dataset across four configurations (Fig.~\ref{fig:loss_dynamics}). 

In the absence of simulated spatial shifts, the full DynoDINO model (Fig.~\ref{fig:loss_dynamics}a) demonstrates smooth, stable loss reduction and consistent Pseudo Dice growth. However, removing the gating mechanism even in this seemingly aligned scenario (Fig.~\ref{fig:loss_dynamics}b) leads to sudden instability; around epoch 200, the validation Pseudo Dice scores become highly erratic and eventually collapse to zero. This unstable behavior confirms that the inherent, sub-voxel misalignments present in clinical post-TACE scans are sufficient to trigger severe representation conflicts in raw subtraction-based architectures when left unmitigated.

Under simulated spatial noise (In-plane Shift), the functional necessity of the Adaptive Gating module is absolute. Without this gating mechanism (Fig.~\ref{fig:loss_dynamics}d), the network suffers from a catastrophic gradient collapse, and the training process fails to initiate entirely, with Pseudo Dice scores remaining at zero throughout the optimization cycle. This suggests that spatial misalignment noise, when amplified by difference-based residual subtraction, completely overwhelms the backpropagated gradient signal. Conversely, the gated DynoDINO model (Fig.~\ref{fig:loss_dynamics}c) successfully filters these spatial artifacts, maintaining stable convergence and robust loss minimization. 

These combined learning dynamics demonstrate that the Adaptive Gating mechanism is not merely an empirical enhancement, but an indispensable architectural ``safety valve.'' It effectively filters representation conflicts and ensures robust optimization in clinical workflows characterized by either inherent scanning misalignments or simulated spatial noise.

\subsection{ZNCC Alignment Visualization}

\begin{figure*}[t]
    \centering
    \begin{subfigure}[b]{0.48\textwidth}
        \centering
        \includegraphics[width=\linewidth]{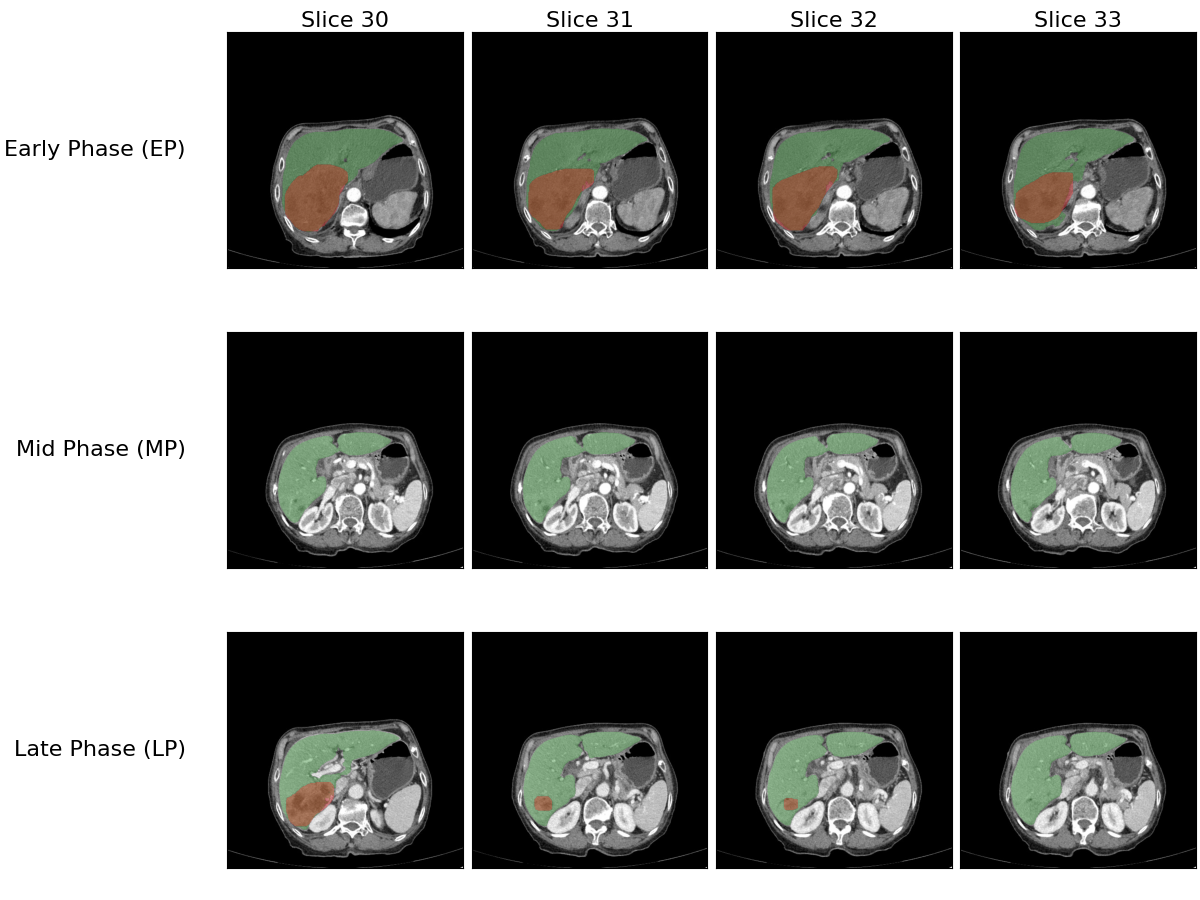}
        \caption{Pre-alignment: Anatomical shifts due to respiratory motion.}
        \label{fig:pre_align}
    \end{subfigure}
    \hfill
    \begin{subfigure}[b]{0.48\textwidth}
        \centering
        \includegraphics[width=\linewidth]{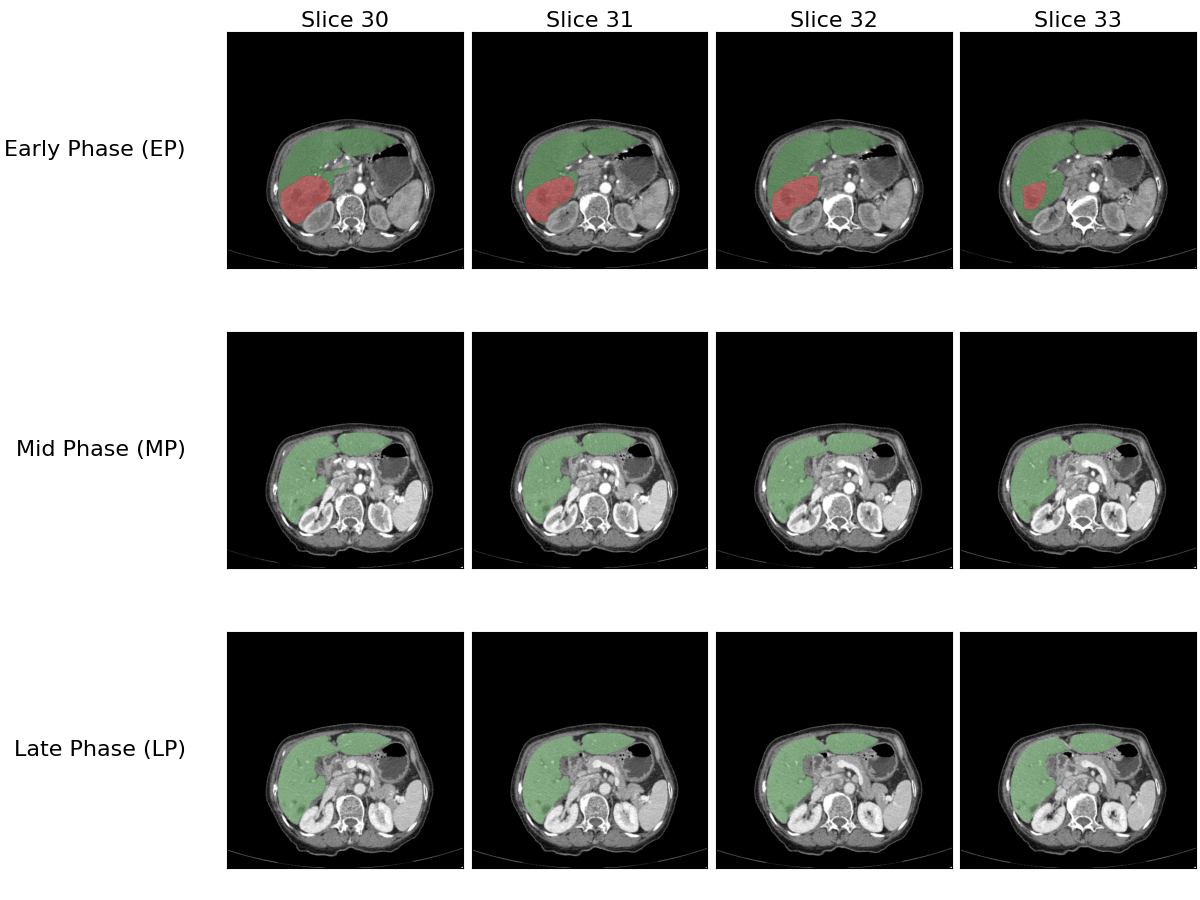}
        \caption{Post-alignment: Spatial synchronization via FFT-ZNCC.}
        \label{fig:post_align}
    \end{subfigure}
    \caption{Visualization of the FFT-ZNCC alignment (Case P0001, PLC-CECT). The green and red masks represent the liver and lesion, respectively. For this specific case, MP is dynamically designated as FP due to its minimum slice count.
    }
    \label{fig:alignment_viz}
\end{figure*}

As visualized in Fig.~\ref{fig:alignment_viz}, FFT-ZNCC alignment ensures that anatomical structures---highlighted by the liver (green) and lesion (red) masks---are spatially synchronized across the EP, MP, and LP. This pixel-wise consistency effectively validates the contribution of this module.


\subsection{Model Complexity and Efficiency Analysis}

The number of trainable parameters and empirical training times for each configuration are summarized in Table~\ref{tab:parameters}. While DynoDINO possesses a higher parameter count (163.85M) compared to the CNN-based baseline nnU-Net (33.47M), this increase is primarily attributed to the high-capacity self-supervised ViT backbone adopted from MedDINOv3 (110.72M).

When extending from single-phase to multi-phase configurations, DynoDINO introduces necessary structural overhead. However, the proposed Mixed-Attention within our inter-phase fusion module significantly mitigates computational complexity. Specifically, the MA mechanism achieves a 4.1\% reduction in the model size (saving 7.08M parameters) compared to a full cross-attention configuration (Table~\ref{tab:ablation}). This architectural efficiency demonstrates that DynoDINO effectively balances high-capacity representation learning with parameter and memory efficiency.


Table~\ref{tab:parameters} reports the empirical wall-clock training times across all three cohorts. Although DynoDINO requires extended training hours due to its multi-phase fusion architecture (20.76--24.25 hours depending on the dataset scale), the computational footprint remains highly manageable for offline clinical training pipelines. This additional training investment represents a highly acceptable trade-off given the substantial gains in morphological fidelity and edge precision across different clinical imaging modalities.

\begin{table}[t]
\centering
\footnotesize
\caption{Comparison of model complexity and training time (notation: LiTS / PLC-CECT / WAW-TACE).}
\label{tab:parameters}
\begin{tabular}{l cc}
\toprule
Method & Trainable Params. & Training Time (h) \\
\midrule
Dino U-Net               & \textbf{13,182,247} & 11.42 / 10.07 / 11.49 \\
nnU-Net                  & 33,473,906          & \textbf{3.69} / \textbf{3.25} / \textbf{3.76} \\
SegFormer                & 27,448,918          & 5.13 / 4.55 / 5.15    \\
MedDINOv3                & 110,718,723         & 7.80 / 7.10 / 7.88    \\
\textbf{DynoDINO (Ours)} & 163,852,803         & 24.25 / 20.76 / 23.99 \\
\bottomrule
\end{tabular}
\end{table}

\section{Conclusions}

This research introduced \textbf{DynoDINO}, a robust framework designed to address the inherent challenges of multi-phase medical image segmentation. By integrating a specialized \textbf{Multi-phase Fusion Model} anchored by LP guidance with the MedDINOv3 backbone, our approach effectively captures complex temporal contrast signatures across diverse clinical scenarios. 

Our experimental analysis demonstrates that the combination of \textbf{Mix-attention} and \textbf{difference-based interaction} allows the model to leverage dynamic contrast-agent kinetics—such as characteristic ``wash-in'' and ``wash-out'' signatures—to resolve spatial ambiguities that are indistinguishable in single-phase snapshots. Significantly, ablation studies validate that the \textbf{Adaptive Gating Mechanism} is not merely an incremental enhancement but a foundational component for architectural stability; it effectively screens out local registration residuals, shielding the transformer backbone from disruptive noise and preventing catastrophic training divergence. While achieving a 4.1\% reduction in total parameter overhead (saving 7.08M parameters), DynoDINO consistently delivers superior boundary precision, as evidenced by substantial gains in \textbf{NSD} and \textbf{HD95} metrics across three large-scale datasets under both aligned and perturbed settings.

While this study focused on liver lesion segmentation as the primary evaluation task, the architectural design of DynoDINO holds the potential to be scaled beyond a single organ or disease. The framework offers a flexible foundation that could be adapted to a broader range of contrast-enhanced imaging applications, including multi-parametric MRI (mpMRI), dynamic perfusion imaging, and other dynamic contrast-enhanced (DCE) protocols \cite{pellicer2022deep, li2024dynamic}. Future work will investigate these clinical contexts to extend DynoDINO's versatility as an efficient solution for high-fidelity multi-phase medical image analysis.

\bibliographystyle{unsrtnat}
\bibliography{references}  






\end{document}